\documentclass[journal]{IEEEtran}
\usepackage[usenames]{color}
\usepackage{hyperref}
\usepackage{jumoline}
\usepackage{ulem}
\usepackage{proofread} 
\noproofreadmark

\usepackage{tipa}

\usepackage{cite}
\let\em\relax
\DeclareTextFontCommand{\em}{\it}
\let\emph\relax
\DeclareTextFontCommand{\emph}{\it}

\usepackage{graphicx}

\usepackage{bm} 
\usepackage{epsfig} 
\usepackage{mathptmx} 
\usepackage{times} 
\usepackage{amsmath} 
\usepackage{amssymb}  
\usepackage{algorithm}
\usepackage{algorithmic}

\title{Symbol Emergence in Cognitive Developmental Systems: a Survey}

\author{%
Tadahiro Taniguchi${}^1$\thanks{${}^1$T. Taniguchi is with College of Information Science and Engineering, Ritsumeikan University, 1-1-1 Noji Higashi, Kusatsu, Shiga 525-8577, Japan {\tt\small  taniguchi@em.ci.ritsumei.ac.jp}}, %
Emre Ugur${}^2$\thanks{${}^2$E. Ugur is with Dept. of Computer Engineering, Bogazici University, Istanbul, Turkey}, %
Matej Hoffmann${}^3$\thanks{${}^3$M. Hoffmann is with Dept. Cybernetics, Faculty of Electrical Engineering, Czech Technical University in Prague, Czech Republic}, %
Lorenzo Jamone${}^4$\thanks{${}^4$L. Jamone is with ARQ (Advanced Robotics at Queen Mary), School of Electronic Engineering and Computer Science, Queen Mary University of London, UK}, %
Takayuki Nagai${}^5$\thanks{${}^5$T. Nagai is with Faculty of Informatics and Engineering, The University of Electro-communications, 
Japan}, %
Benjamin Rosman${}^6$\thanks{${}^6$B. Rosman is with Mobile Intelligent Autonomous Systems, at the Council for Scientific and Industrial Research, and with the School of Computer Science and Applied Mathematics, at the University of the Witwatersrand, South Africa}, %
Toshihiko Matsuka${}^7$\thanks{${}^7$T. Matsuka is with Department of Cognitive and Information Science, Chiba University. 
Japan}, %
Naoto Iwahashi${}^8$\thanks{${}^8$N. Iwahashi is with Faculty of Computer Science and Systems Engineering, Okayama Prefectural University, 
, Japan}, %
Erhan Oztop${}^9$\thanks{${}^9$E. Oztop is with Dept. of Computer Science, Ozyegin University, Istanbul, Turkey}, %
Justus Piater${}^{10}$\thanks{${}^{10}$J. Piater is with the Dept.\ of Computer Science, Universit\"at Innsbruck, Austria.}, %
Florentin W\"org\"otter${}^{11}$\thanks{${}^{11}$F. W\"org\"otter is with the Department for Computational Neuroscience at the Bernstein Center of the University of G\"ottingen, Inst. f. Physics III, Germany.}
}

\begin{document}
\maketitle

\begin{abstract}
\addspan{Humans use signs, e.g., sentences in a spoken language, for communication and thought. Hence, symbol systems like language are crucial for our communication with other agents and adaptation to our real-world environment. The symbol systems we use in our human society adaptively and dynamically change over time. In the context of artificial intelligence (AI) and cognitive systems, the symbol grounding problem has been regarded as one of the central problems related to {\it symbols}. However, the symbol grounding problem was originally posed to connect symbolic AI and sensorimotor information and did not consider many interdisciplinary phenomena in human communication and dynamic symbol systems in our society, which semiotics considered. In this paper, we focus on the symbol emergence problem, addressing not only cognitive dynamics but also the dynamics of symbol systems in society, rather than the symbol grounding problem. We first introduce the notion of a symbol in semiotics from the humanities, to leave the very narrow idea of symbols in symbolic AI. Furthermore, over the years, it became more and more clear that symbol emergence has to be regarded as a multifaceted problem. Therefore, secondly, we review the history of the symbol emergence problem in different fields, including both biological and artificial systems, showing their mutual relations. We summarize the discussion and provide an integrative viewpoint and comprehensive overview of symbol emergence in cognitive systems. Additionally, we describe the challenges facing the creation of cognitive systems that can be part of symbol emergence systems.}
\end{abstract}


\IEEEpeerreviewmaketitle

\begingroup
\renewcommand{\thefootnote}{}
\footnotetext{This research was partially supported by Grants-in-Aid for Young Scientists (A) (15H05319) and for Scientific Research on Innovative Areas (16H06569) funded by the Ministry of Education, Culture, Sports, Science, and Technology, Japan and by CREST, JST. Matej Hoffmann was supported by the Czech Science Foundation under Project GA17-15697Y. F. W\"org\"otter has been funded by the European H2020 Programme under grant 732266, Plan4Act. Emre Ugur and Justus Piater have received funding from the European Union’s H2020 research and innovation programme under grant agreement no. 731761, IMAGINE. Our thanks go to Dr. M. Tamosiunaite and Mr. Hamit Basgol for their critical comments on the manuscript.}
\endgroup

\section{INTRODUCTION} \label{sec:Introduction}

\begin{figure}
  \begin{center}
\includegraphics[width=80mm]{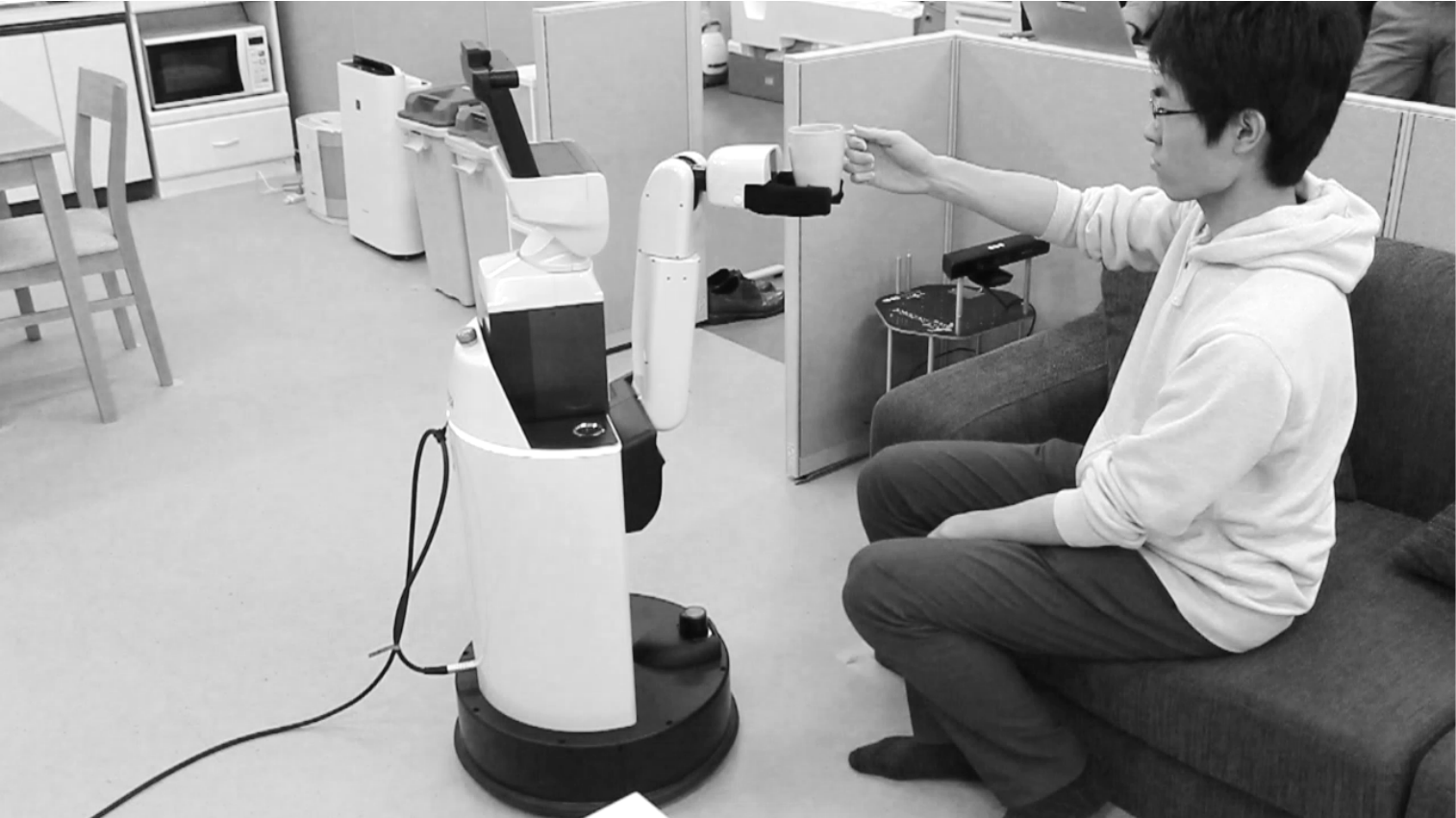}
\caption{Robot in a home environment that has to deal with complex manipulation, planning, and interaction via semiotic communication with human users.}
\label{fig:robot}
\end{center}
\end{figure}
\addspan{\IEEEPARstart{S}YMBOLS, such as language, are used externally and internally when we communicate with others, including cognitive robots, or think about something. 
Gaining the ability to use symbol systems, which include not only natural language but also gestures, traffic signs, and other culturally or habitually determined signs, in a bottom-up manner is crucial for future cognitive robots to communicate and collaborate with people. Because symbols in the human society are not given by a designer but instead they emerge and dynamically evolve through social interaction, the problem of symbol emergence in cognitive and developmental systems is central to advance the state of the art of intelligent human-machine interaction.}

\addspan{Figure~\ref{fig:robot} depicts a human--robot interaction scenario. Here, a robot is handing over a cup following the utterance of the command ``give me the cup" by the person. A string of letters (written or spoken) has to be translated into a series of motor actions by the robot, with reference to the context and the environment. This may appear to be a simple task. However, to achieve this \emph{in a developmental manner}, the machine has to learn to deal with complex manipulation, planning, and interaction via natural semiotic communication, i.e., communication using symbols, with the human. Additionally, semiotic communication in the real-world environment usually relies on mutual beliefs, shared knowledge, and context between the speaker and listener to resolve uncertainties in the communication and environment.
Ultimately, this amounts to providing the robot with a set of mechanisms for developing a complex cognitive system. The mechanisms should allow the robot to learn representations and symbols for semiotic communication. 
Thus, we require powerful machine-learning methods that emulate appropriate models of the cognitive development of this human trait. Such a set of methods must start with a hierarchical decomposition of spatiotemporal and sensorimotor data fusing multimodal information. The main challenge is then the autonomous generation of a wide range of grounded concepts that are internally represented with different levels of abstraction. This will allow the agent to make inferences across different complexity levels, thereby allowing it to understand this type of human request.}

\addspan{In addition to learning or internalizing the pre-existing meaning and usage of signs, we humans can also invent and generate signs to represent things in our daily life. This immediately suggests that symbol systems are not static systems in our society, but rather dynamic systems, owing to the semiotic adaptiveness and creativity of our cognition. However, current robots still do not have sufficient symbol system learning capabilities to become able to communicate with human users or to adapt to the real-world environment in a developmental manner.}

\addspan{With respect to the term ``symbol'', many scholars still have significant confusion about its meaning. They tend to confuse symbols in symbolic AI with symbols in human society. When we talk about symbol emergence, we refer to the latter case. Note that this paper is not about symbolic AI, but rather focuses on the question of how we can develop cognitive systems that can learn and use symbol systems in our society.}

\addspan{The symbol grounding problem was proposed by Harnad~\cite{Harnad1990}. Their paper started with symbolic AI. Symbolic AI is a design philosophy of AI based on the bold physical symbol system hypothesis proposed by Newell~\cite{Newell1976,Newell1980} (see Section~\ref{sec:hist_ai}), which has already been rejected practically, at least in the context of creating cognitive systems in the real-world environment. 
The term ``symbol'' in symbolic AI is historically rooted in symbolic/mathematical logic. It is originally different from the symbols in our daily life. 
The motivation of Harnad's study was to determine how to connect symbolic AI and sensorimotor information to overcome the problem that such symbols, i.e., tokens, have no connections with our real-world phenomena without grounding. Since that paper was published, many scholars have presented broad interpretations of the symbol grounding problem, and the problem definition has become more and more popular. However, because the discussion of the symbol grounding problem started with a questionable assumption, i.e., using symbolic AI, the problem formulation of the symbol grounding problem has several drawbacks.}

\addspan{First, it is therein assumed that symbols are physical tokens and considered that symbols in communication (e.g., words), and tokens in one’s mind, (i.e., internal representations) are the same. This comes from the physical symbol system hypothesis, which we now need to say is simply wrong. It is fair to assume that our brain does not contain such strictly discrete internal representations. Instead, it is more likely that there exist more probabilistic, continuous, and distributed representations, e.g., activation patterns in neural networks and probability distributions in probabilistic models, which lead to computational requirements different from those in (classical) AI. Second, their classical paper did not consider the social dynamics and diversity of symbol systems, which are crucial in semiotics. In other words, the meaning of a symbol can change over time and between different communities and contexts.}

\addspan{Meanwhile, we still have challenges in making artificial cognitive systems that can learn and understand the meaning of symbols, e.g., language. However, most studies, tackling the challenges in developing cognitive systems that can learn language and adaptive behaviors in the real-world environment, rarely use symbolic AI nowadays. Cognitive systems do not have to have pre-existing symbol systems for their cognition. Instead, an agent ought to learn a symbol system, e.g., language, by interacting with others and its environment, and even invent new symbols and share them with the community it joins. Therefore, naturally, symbol grounding problem is, under this viewpoint, replaced with the symbol emergence problem. Symbol emergence has been suggested as a critical concept for understanding and creating cognitive developmental systems that are capable of behaving adaptively in the real world and that can communicate with people~\cite{Plunkett1992,Taniguchi2016SER}.}

\addspan{Thus, the symbol emergence problem is indeed a multifaceted problem that should take different sources of information for learning into account. The required computational models should enable the simultaneous learning of action primitives together with the syntax of object manipulations, complemented by lexicographic, syntactic, and ontologic language information learned from speech signals. To scale up to real-world problems, the required architecture needs to integrate these aspects using general, not specifically tailored, mechanisms that provide an overarching framework for cognitive development.
}

\addspan{Thus, in this paper, we address the interdisciplinary problem history of symbols and symbol emergence in cognitive developmental systems discussed in different fields and also point out their mutual relations}\footnote{This paper is based on a one-week workshop, the Shonan meeting on \emph{Cognitive Development and Symbol Emergence in Humans and Robots} (\url{http://shonan.nii.ac.jp/shonan/blog/2015/10/31/cognitive-development-and-symbol-emergence-in-humans-and-robots/}). This meeting focused on a constructive approach toward language acquisition, symbol emergence, and cognitive development in autonomous systems, discussing these from highly interdisciplinary perspectives.}\addspan{. We describe recent work and approaches to solve this problem, and provide a discussion of the definitions and relations among related notions, such as \textit{concept}, \textit{category}, \textit{feature representation}, and \textit{symbol}. These terms have often been used interchangeably in different theories and fields to refer to similar phenomena. This paper aims to provide a comprehensive survey of studies related to symbol emergence in cognitive developmental systems~\cite{Asada2009}, as well as an integrative viewpoint on symbol emergence describing potential future challenges.}

Section~\ref{sec:hist_semi} introduces the aspect of semiotics.
Sections \ref{sec:ProblemHistory1} and \ref{sec:ProblemHistory2} describe the problem history in related fields concerning biological systems (i.e., mainly humans) and artificial systems, respectively. Section~\ref{sec:IntegrativeView} provides an integrative viewpoint that is consistent with the notion of symbols in the different fields. 
Section~\ref{sec:Challenges} describes the challenges in this type of research, and Section~\ref{sec:Conclusion} offers some conclusions.

\begin{figure}
\begin{center}
\includegraphics[width=80mm]{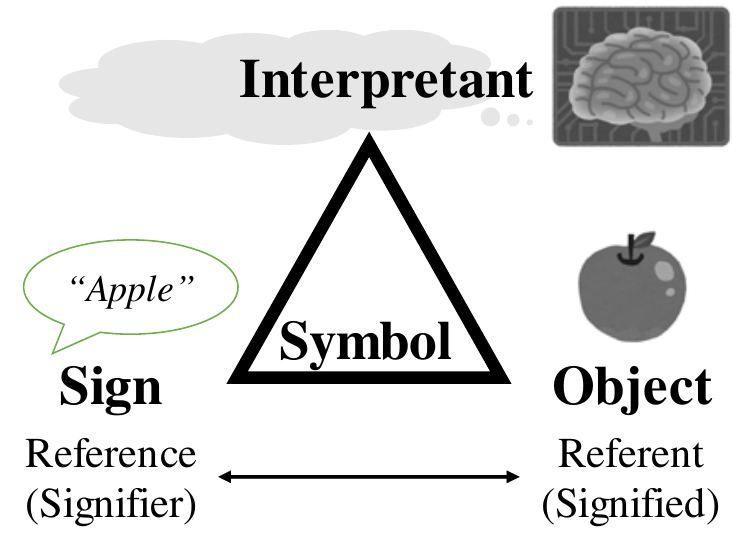}
\caption{Definition of a symbol in semiotics. The triadic relationship between the sign, object, and interpretant is regarded as a symbol. In a static case, it tends to be regarded as a dyadic relationship between the reference and referent. Note that in general, the interpretation of signs also depends on the context.}
\label{fig:semiosis}
\end{center}
\end{figure}

\section{Semiotics: from signs to symbols}\label{sec:hist_semi}

\textit{Symbol} is not only a term in AI and cognitive robotics, but is also frequently used in our human society. The nature of the symbols that humans use in their daily life is studied in {\it semiotics}, which is the study of {\it signs} that mediate the process of meaning formation and allow for meaningful communication between people. This is a highly interdisciplinary research field related to linguistics, cultural studies, arts, and biology. Peirce and Saussure are considered as independent founders of this field~\cite{Chandler2002}. 

Endowing robots with the capability to understand and deal with symbols in human society (e.g.,\ Fig.~\ref{fig:robot}) requires a robot to learn to use symbol systems that we humans use in our society. Symbols are formed, updated, and modulated by humans to communicate and collaborate with each other in our daily life. Therefore, any definition of \textit{symbol} that \textit{only} holds for artificial systems is insufficient to allow for human--robot interaction. 

To achieve this, we can adopt the definition of \textit{symbol} given by Peircean semiotics and represented by a semiotic triad (Fig.~\ref{fig:semiosis}). Peircean semiotics considers a symbol as a process called {\it semiosis}. Semiosis has three elements, i.e., {\it sign (representamen)}, {\it object}, and {\it interpretant}. The triadic relationship is illustrated in Fig.~\ref{fig:semiosis}. \textit{Sign} describes the form that the symbol takes, e.g., signals, utterances, images, facial expressions, etc. \textit{Object} refers to something that the sign represents. \textit{Interpretant} (rather than interpreter) is the effect of a sign on a person who perceives the sign. In other words, the \textit{interpretant} is the process that relates the \textit{sign} with \textit{object}.

Naively, people tend to think there is a fixed relationship between a sign and its object. In our natural communication and social life, the meaning of signs hugely depends on contexts, participants, communities, cultures, and languages. It is the third term---interpretant---which gives our semiotic communication a very high degree of diversity. A symbol is not a static entity, but rather a dynamic process involving active interpretation. The dynamic inference process is named semiosis in Peircean semiotics~\cite{Peirce}. Thus, the meaning that a sign conveys to a person can change depending on its interpretant. 

In this paper, we employ this definition of \textit{symbol} because our target is semiotic communication not symbolic AI. Note that signs do not have to be speech signals or written letters. Any type of stimulus, i.e.\ any signal, could become a sign in a symbol. 

By contrast, Saussurian semiotics\footnote{Saussurian semiotics is often called semiology as well.} places more emphasis on a systematic perspective of symbols as opposed to the individual dynamics of a symbol. This model assumes that \textit{symbol} consists of \textit{sign} (signifier) and \textit{referent} (signified) leading to a dyadic structure. This view is similar to the view of symbolic AI. In symbolic AI, to determine the meaning, more emphasis is placed on symbol--symbol relations.

Common to both---Peircean and Saussurian semiotics---is the fact that \textit{sign} and \textit{object} are heavily intertwined. A sign that is not linked to objects or events directly or indirectly is essentially meaningless~\cite{Harnad1990}. Only when we perceive the sign of a symbol, can we infer which object the sign represents. Furthermore, it should be noted that a symbol generates a boundary between the ``signified'' (object) and anything else. Therefore, symbol systems introduce discrete structures into the world.

The supporters of both models also insist that the relationship between \textit{sign} and \textit{referent} (object) is arbitrary. Arbitrariness does not indicate randomness or disorder. Rather, it means that symbol systems defining relationships between signs and referents can be different from each other, depending on languages, cultures, communities, and generations. This is an important aspect of symbol systems in our human society.

Scholars who may be familiar with symbolic AI may argue that the advantage of symbolic AI is that one can manipulate symbols independently of the substrate on which they are grounded. Arguably, this has been regarded as the central, major strength that makes symbol systems so useful. \addspan{As remarked by Harnad~\cite{Harnad1990}, a symbol is a part of a symbol system, i.e., the notion of a symbol in isolation is not a useful one.} The question about symbol manipulation has also led to a vigorous debate between the Cartesian view (mind without body) of symbolic AI as compared to the view presented by \emph{embodied cognition}, which posits that the mind without the body is impossible~\cite{Wilson2002,Woergoetter2009,PfeiferBongard2007}. 

Numerous studies in semiotics, including cultural studies, provided a deeper understanding of symbol systems. However, so far there is no computational model that can reproduce the dynamics of semiosis in the real world. In addition, most of the studies in semiotics focused on existing symbol systems and rarely considered the developmental process and emergent properties of symbol systems.
\addspan{Presuming the definition of symbols in semiotics, rather than starting with symbolic AI, we would like to integrate interdisciplinary views of symbol systems, and set out an integrative common foundation called the {\it symbol emergence system}.}

\section{Problem history 1: Symbol emergence in biological systems}\label{sec:ProblemHistory1}

\subsection{Evolutionary viewpoint: from actions to symbols}\label{sec:hist_evo}

\begin{figure}
\begin{center}
\includegraphics[width=90mm]{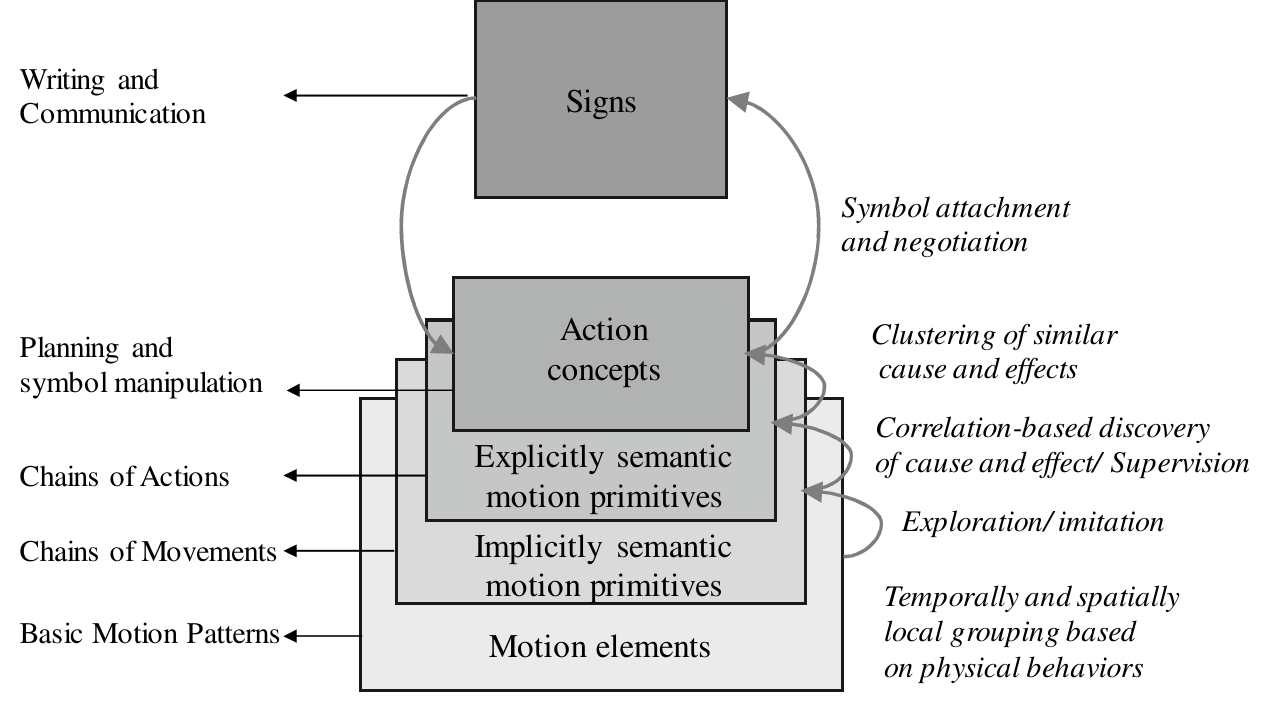}
\caption{Pyramid of cognitive complexity leading upward from basic motor control to symbolic representations.}
\label{fig:action}
\end{center}
\end{figure}
\addspan{From an evolutionary viewpoint, symbol emergence should also be explained in the context of adaptation to the environment. }
\delspan{Which processes are needed to lead to symbol emergence? Is it the aspect of attaching a \textit{sign} to an \textit{object} (dyadic model!) that poses the biggest problem to a cognitive agent, or what else makes symbol emergence difficult? In the first place we would argue that attaching signs to objects is not very hard. Rather it is the question: ``What does \textit{object} mean?'' that poses a difficult problem as shall be discussed next.}
\delspan{Symbols are at the far end of a whole chain of complex processes, and many animals can perform earlier steps in this chain but not later ones, making them ``less cognitive'' than humans. What do we have in mind here? Fig.~\ref{fig:action} depicts a possible pyramid that leads  from simple, clearly non-cognitive processes step by step upwards to complex, symbol-supported (language-supported) cognition. Essentially we posit that such a chain moves from combination of motion elements via semantic motion primitives to the formation of concepts where signs are attached (are created) only at the very end.}
With few exceptions, and different from plants, all animals are able to move. Considering the phylogeny of the animal kingdom, movement patterns, actions, and action sequences have become more and more complex. In parallel to this development, more complex cognitive traits have emerged over 2--3 billion years from protozoa to humans, too. Actions, action sequences, and action planning have been discussed as structures highly similar to grammar~\cite{Greenfield1991,Fujita2009}. For example, Aloimonos et al.\ and Lee et al.\ have developed a context-free action grammar to describe human manipulations~\cite{pastra11grammar,Lee2015}.

We would now argue that \textit{action} and \textit{action understanding} come before symbolization, and that the grammatical structure implicit to action lends itself as a natural scaffold to the latter. 

The vast majority of all animals perform their movements almost exclusively in a reactive, reflex-like manner, responding to stimuli from the environment. This leads to a repertoire of various movement patterns, which we could consider as the basic syntactic elements in a grammar of actions. Chaining such actions leads to action sequences. A jellyfish can do this: on touching, poisonous microharpoons are triggered in its tentacles, after which the victim is hauled in and finally devoured. Remarkably, such an action chain has deep semantics (meaning) in terms of survival for the jellyfish; still, we would think it is fair to assume that it has no inkling about its own actions. Notably, evolution has built many highly ``programmed'' and reactive systems (e.g., spiders) that can chain actions in very complex ways (e.g. building their net). Semantics, while present, nevertheless remain fully implicit to all of those.

The making-explicit of action semantics does not come easily, and action understanding is not a one-shot process that suddenly appeared in evolution. Potentially, all those animals that can purposefully ``make things happen'' achieve first steps along this path. In other words, a creature that is able to prepare the grounds for a successful action by making sure that all necessary action preconditions exist has already gained some action understanding. This creature would not yet be able to reason about its actions, but it can purposefully repeat them. The more flexibility such a creature shows in being able to perform an action under different and variable situations, the higher we would deem its cognitive standing. Somewhere, our imagined animal has begun to understand the cause--effect relations that underlie its actions. Correlation-based learning, possibly paired with imitation learning, can underlie the formation of this type of action understanding in these animals\footnote{Here, supervised learning plays a major role for humans, also found in great apes to a lesser degree.}. The fact that this type of learning is only associative and not inferential (as discussed in comparative psychology~\cite{penn07causalCog,hanus16causalCog}) may be less relevant to our practical stance: if I can purposefully make it happen, I have understood it.

Repetitions of the same or a similar action in different situations, and hence of different action variants in an action class, can make it possible to extract action pre- as well as post-conditions, the cause and effect structure, or in one word: the \textit{essence} of this action class. This requires memory as well as generalization processes that allow commonalities from individual examples to be extracted. In the regime of machine learning, clustering mechanisms across individual instantiations of action-triggered cause--effect pairs could possibly serve this purpose. For humans and a few ``higher'' animals, something similar is happening and---if successful---one could say that this animal/human has formed a \textit{concept} of the action. The term \textit{concept} refers here to an overarching structure because it subsumes all essential pre-conditions, action variants, and action outcomes (post-conditions) into one conjoint representation, omitting all aspects that are noisy and/or contingent. This process also leads to the fact that concepts are discrete, making them distinguishable from each other.

Symbols are at the far end of a whole chain of complex processes, and many animals can perform earlier steps in this chain but not later ones, making them ``less cognitive'' than humans. What do we have in mind here? Figure~\ref{fig:action} depicts a possible pyramid that leads from simple, clearly noncognitive processes step by step upward to complex, symbol-supported (language-supported) cognition. Essentially, we posit that such a chain moves from a combination of motion elements via semantic motion primitives to the formation of concepts where signs are attached (created) only at the very end.

Whereas action may have been the origin for triggering concept formation, by now we---humans---have surpassed this stage and have taken concept formation into different and sometimes highly abstract domains. 

We would argue that the formation of such discrete concepts is central to the emergence of symbols and that the level of concepts is the one where symbolization can set in. The term \textit{Object} in Fig.~\ref{fig:semiosis} indeed takes the wider meaning of \emph{concept of object (-class)}.

In particular, one can now see that attaching a sign to a concept is simple. You could use anything.
Is this all and are we done now?

We believe the remaining difficulty lies in the communication process. Symbols (based on signs) are initially \textit{only} meaningful in a communication process. This may have started as one-step memos of an individual, such as the drawing of a wiggly line on the ground to remind your stone-age self to go to the river as suggested by Dennett~\cite{Dennett1991}. This may have led to more complex silent discourse with yourself and in parallel to true inter-individual communication. It is indeed totally trivial to attach a sign, say a string of letters, to a concept (once the latter exists), but to understand my uttering of this string, my listener and I need to \textit{share the same or at least a very similar concept} of the referred-to entity. Thus, we would argue that the forming of a concept is a difficult problem and so is the negotiation process for the sharing of a symbol (where the actual shape of the sign does not matter). 

Now, it also becomes clearer why humans and some nonhuman primates can handle symbols\footnote{Bonobos can handle but not create symbols. The latter has, at best, been reported by anecdotal observations.}: These genera are highly social and continuously engage in inter-individual exchanges; they communicate. Wolves do this, too. Why did they not begin to develop symbolic thinking? We believe this is where the grammatical aspect of action comes in. Action \textit{chains} lend themselves to planning and reasoning about them but only if these chains are performed in a highly reliable way, for example by the hands of monkeys (humans). Wolves cannot do this. They cannot reliably chain (manipulate) actions. The \textit{reliable} repeatability of action chains may first lead to planning and, second, to explicit symbolization, which would be the next step.

While suggestive, it necessarily remains unclear whether or not evolution indeed followed this path. We are here, however, centrally concerned not with this but rather with artificial cognitive developmental systems, asking how they could learn to use symbols. The above discussion posits that this could possibly be best understood by tackling the concept formation and symbol negotiation problem.  

\subsection{Neuroscientific viewpoint: from neural representations to symbols}\label{sec:hist_neuro}

Humans use symbols. 
\delspan{Employing terminology from computer science, it is fair to say that---indeed---we compute, perform message passing, interrupt handling, memory storage, retrieval and other similar symbol-based operations. Do our brains use symbols, too? This sounds like a strange question but it can be rephrased like this: Are there discrete entities present inside the brain on which the brain performs similar operations? And would these entities concur with a Peircean (or Saussurian) model for \textit{symbol}?}
On the one hand, it is known that all the representations in the brain are highly dynamic and distributed, and thus only a snapshot of the brain's activity and state could come close to what we may call a symbol in symbolic AI. Such a fine-grained view would, however, be useless as it completely spoils the utility of symbol systems that compactly represent concepts (things, ideas, movement patterns, etc.). On the other hand, we could hope to find local neural representations that stand for abstractions of percepts, events, motor plans, etc., which could thus be considered concepts of the brain.

This view critically assumes that concept-like internal representations in the brain must have emerged before expressed symbols for social communication. The initial evolutionary pressure for their formation was not social communication but rather for internal computation and communication in the brain. 
\delspan{Surely, from an evolutionary viewpoint such a neural system prepares the ground for symbol/concept externalization, and the start of language when combined with recursive structure of action planning and imitation as elaborated in Arbib's language evolution theory. }
There are intriguing examples of such localized neural representations that may play the role of a concept; however, whether these symbol- or concept-representations are manipulated for inference and planning is largely unknown. 

A good candidate for the representation of action concepts is the neural response exhibited by {\it mirror neurons}. These neurons were initially discovered in the ventral premotor cortex of macaque monkeys (area F5), and evidence exists that humans are also endowed with a mirror system\cite{RN407}. Macaque mirror neurons become activated when a monkey executes a grasp action, as well as when the monkey observes another monkey or human perform a similar action~\cite{RN37,RN34,RN35}. This duality lends support to the plausibility of the idea that a mirror neuron activity represents an abstraction of an action for further processing, as opposed to being an intermediate representation of action production. If the former is indeed correct, then mirror neurons should be considered as encoding concepts or symbols for actions, regardless whether they are executed or observed. 

In general, the minute details of an action are not critical for most of the mirror neurons\cite{RN27}; therefore, their activity is probably not a precise motor code. Initial reports on mirror neurons focused on their role for action understanding; but later, higher-level functions have also been attributed to them, such as understanding the intention of the observed actions (e.g.,\ \emph{grasping for eating} vs.\ \emph{grasping for placing})~\cite{RN475}. 

It would be very valuable to decipher how mirror neuron activity is used elsewhere in the brain. If it were used as the basis for planning, we could happily claim that some sort of concept representation is undertaken by these neurons. In this case, we may feel comfortable calling the mirror activity a neural symbol of an action.

Human brain imaging studies suggest that significant conceptual binding of objects, events, actions, feelings etc.\ exists in different mirror systems. For example, feeling disgust and seeing someone disgusted activates overlapping regions in the brain~\cite{RN978}. This is also the case for pain~\cite{RN672}. It would, therefore, be possible to claim that these overlapping areas represent the concept of pain, and the neural (population) activity pattern there could be considered as the neural symbol representation for pain. Opposing this view, however, is that these are gross symbols that lack manipulability, and thus cannot really fulfill the role of \textit{symbol} as defined above.  
\delspan{We probably consider digits as one of the most representative symbols of all. Digits indicate the cardinality  of sets regardless of the identity of the elements and how we sense them. Growing evidence in neuroscience indicates that the primate brain does have its ``neural digits'', i.e.\ a local neural pattern of activity in the parietal cortex that represents numerosity of a set irrespective of the identity of the elements, and for some neurons irrespective of the sensory domain by which the elements are sensed  (an extensive review can be found in). }
\addspan{From an evolutionary viewpoint such a neural system prepares the ground for symbol/concept externalization, and the start of language when combined with the recursive structure of action planning and imitation as elaborated in Arbib's language evolution theory~\cite{RN344,RN977,RN353}. 
Instead of the evolution of a syntax processing enabled brain related to speech production, Arbib argues that the neurons for manual actions have evolved into mirror systems for imitation and simple manual communication, which facilitated the biological evolution of the human language-ready brain that can sustain a protolanguage~\cite{RN977,rizzolatti1998language}. The transition from protolanguage to language is then explained by a long history of human cultural evolution.  The theory is supported by the findings that overlapping brain activity is observed in Broca’s for perceiving language and tool use~\cite{higuchi2009shared}, both of which involve  manipulation of complex hierarchical structures. Furthermore, as Broca’s area is the possible homologue of the monkey ventral premotor cortex where mirror neurons for grasping are located, neural processes for manipulation and action understanding may be the evolutionary bases of language~\cite{arbib2011mirror}.}

\addspan{The crucial finding, refered to in this theory, was that the mirror system activity in the human frontal cortex was found to be in or near Broca’s area, which had been considered as an area of language processing. They hypothesized that changes in the brain that provide the substrate for tool use and language and that allow us to cumulate innovations through cultural evolution are the key innovations in language evolution. The ability to make and recognize praxic actions may have provided the basis for the ability to make and recognize communicative manual gestures, with this in turn providing the evolutionary basis for brain mechanisms supporting the parity property of language: In other words, the evolution of language correlated with the development of mirror neurons that enabled humans to perform manipulation of physical objects, and gestures for communication and language use~\cite{tomasello2010}.}

Therefore, it appears that in highly evolved animals, brains have ``invented'' basic symbols that may be used in rudimentary planning and decision making. In pre-humans, these basic symbols might have triggered an accelerated evolution once they were externalized via utterances, paving the way for speech and language evolution.

\subsection{Cognitive science viewpoint: from concepts to symbols}\label{sec:hist_cog}

Cognitive scientists have been considering symbol systems for many decades~\cite{Newell1980}. Similar to our discussion above (Section~\ref{sec:hist_evo}), most cognitive scientists and psychologists agree that the notion of a {\it concept} is very important in understanding symbols as well as cognition, as it enables us to classify, understand, predict, reason, and communicate~\cite{medin1992cognitive}. 

However, researchers do not necessarily agree on the definition of \textit{concept}. Furthermore, there is a related term, {\it category} and its cognitive competence called {\it categorization}, which needs to be distinguished from \textit{concept}.

The following may help. When we saw a dog, we usually think ``we saw a dog,'' but not ``we saw a hairy living organism with two eyes, four legs, pointy teeth, etc.'' That is, instead of simultaneously interpreting the abundant individual features of an entity, we usually categorize the entity into one class (or a few classes). Thus, categorization can be seen as an effective data encoding process that compresses the information of many features into a few classes. Likewise, using a category can be interpreted as a data decoding process that extracts principal information (e.g., variances and correlations) about a set of features that the category has.  

Let us now try to distinguish concept from category. One of the most accepted definitions of \textit{concept} is that it corresponds to an internal (memory) representation of classes of things~\cite{medin1992cognitive,murphy2004big,margolis2007ontology}. Different from this, one of the most accepted definitions of \textit{category} is that of a set of entities or examples selected (picked out) by the belonging concept~\cite{medin1992cognitive,murphy2004big}. 

Clearly, those two terms are closely related and, thus, often used interchangeably in cognitive science and psychology, leading to controversies. For example, one notable source of disagreement concerns the internal representation asking about possible systems of symbols on which the internal representation of concept could be built. This mainly concerns the difference between an amodal symbol system (ASS) as compared to a perceptual symbol system (PSS)~\cite{margolis2007ontology,barsalou2005situated}.  

A symbol system that uses symbols independently of their (perceptual) origin as atoms of knowledge is called an amodal symbol system (ASS). However, Barsalou more strongly concentrated on the perceptual aspects of symbol systems and proposed the PSS ~\cite{Barsalou1999}. A PSS puts a central emphasis on multimodal perceptual experiences as the basis of its symbols.

Let us describe the differences between ASS and PSS in more detail.  

First, the notion of category is different in ASS and PSS. ASS assumes that categories are internally represented by symbols without any modality, where \textit{symbol} means a discrete variable, which is often given by a word (a name of a concept). By contrast, in PSS, every category is internally represented by symbols from several different modalities originating in the sensorimotor system and resulting in distributed and overlapping representations. This difference is crucial, because the symbol grounding problem remains in ASS, whereas symbols in PSS are directly related to perceptual experiences and, thus, the meanings of symbols are well grounded~\cite{Harnad1990}. 

PSS furthermore assumes that categories are (implicitly or explicitly) consolidated by simulating several candidate categories. For this, distributed and complexly connected perceptual symbols with background information (i.e., context or situations) are used. This suggests that categorization is a dynamic process and is highly influenced by the context and situation. From the viewpoint of PSS, categorization is, thus, situated and flexibly \emph{bound to the world}. This allows humans to be highly adaptive and to exhibit intelligence across several domains.

Second, how concepts are used to realize categories is also different. ASS assumes that a concept is acquired knowledge that resides in long-term memory, and, thus, concepts are more or less static. On the contrary, PSS assumes that concepts are dynamic. Note that in this way, there is a certain similarity between PSS and Peircean semiotics.

There is an interesting phenomenon called {\it ad hoc category}. An ad hoc category is a spontaneously created category to achieve a given goal within a certain context~\cite{barsalou1983ad}. For example, ``things to take on a camping trip'' is an ad hoc category. Ad hoc categories are shown to have the same typical effects seen in more conventional categories such as dogs and furniture. This somewhat odd but interesting cognitive phenomenon can be well described by dynamic simulation processes that utilize situation-dependent, multimodal perceptual symbols in PSS.  
Different from this, ASS cannot account for ad hoc categories because of its static and situation-insensitive nature.  

To date, the ASS perspective still dominates in cognitive science and psychology, probably because it has been more manageable to conduct research assuming static ASS-like representations. PSS, however, seems to be better matched to the dynamics of brain structures and activities, and to robots that need to deal with real-world sensorimotor information. It offers a more natural, distributed representation compatible with modern theories of the dynamic restructuring of neuronal cell assemblies reflecting various memory contents~\cite{Palm_2014,Holtmaat_2016}. So far, however, it has been quite difficult to empirically evaluate the PSS theory as a whole. Potentially, building better computational models for symbol emergence in cognitive developmental systems might aid in making progress concerning the ASS--PSS controversy in cognitive science. This review certainly leans more toward PSS than ASS.

\subsection{Developmental psychology viewpoint: from behaviors to symbols}\label{sec:hist_dev}
The idea that symbols and symbol manipulation emerge from infants' physical and social interactions with the world, prior to their first words and linguistic activity, has been actively discussed since the times of the pioneers of developmental psychology. 

\subsubsection{\addspan{Theories of symbol development}}\addspan{The theories of symbol emergence and development in developmental psychology can be separated into two
categories: classical and contemporary. Classical accounts were developed by
Piaget, Vygotsky, and Werner and Kaplan, and contemporary accounts can be
separated into three main approaches: empiricist, nativist, and constructivist.
For classical theories, the emergence of symbolic representation is an active
process and all authors emphasize the importance of social interaction. However,
the proposed mechanisms were different.}
Piaget's book \emph{La formation du symbol}~\cite{Piaget1962} as well as the book of Werner and Kaplan entitled \emph{Symbol formation}~\cite{werner1963} already discussed the gradual emergence of symbols from nonlinguistic forms of symbol-like functioning, and the integration of the formed structures into language skills. 
\addspan{In contemporary accounts, empiricist
and nativist theories consider a child to be a passive container, whereas
constructivist theories see a child as an active agent trying to construct
meaning through social interaction and communication \cite{Callaghan2015}. 
According to Mandler, the traditional view of infants’ inability of
concept formation and use is flawed. Her studies on recall, conceptual
categorization, and inductive generalization have shown that infants can
form preverbal concepts. An infant can form animals and inanimate
distinction from only motion and spatial information. Mandler argues
that basic preverbal concepts are derived from the spatial information
and suggests that perceptual meaning analysis creates preverbal concepts
with the form of image-schemas~\cite{Mandler2007}.}

\addspan{The levels of consciousness model developed by Zelazo consists of four levels~\cite{Carlson2009}. The first level is called stimulus bound, ranging from birth to 7
months of age, in which an infant has representations tied to stimuli. In
the second level, which is decoupling of symbols, ranging between 8 and 18
months, infants can substitute symbols for stimuli in their working
memory. The third level is named symbols as symbols, in which children
from 18 months to five years of age show genuine symbolic thought and can
engage in pretend play. Starting from five years of age, the fourth level is
quality of symbol-referent relations. In this stage, children can assess
the quality of the relationship between symbol and referent; and finally
can understand ambiguity, sarcasm, artistic representation, and reasoning
scientifically~\cite{Carlson2009}. For Tomasello, symbolic communication emerges as a
way of manipulating the attention of others to a feature of an object or an
object in general. Up to 9 months of age, behaviors of children are
dyadic. Within the age range of 9--12 months, new behaviors emerge that include
joint attention and triadic relationship, i.e., self, other and object.  This is the first time when an infant has the
ability of joint engagement, social referencing, and imitation. He argues
that for infants to understand or engage in symbolic convention, they should see
others’ as intentional agents with a goal \cite{Tomasello2003}. This view reminds us
that the social abilities of an agent would also be critical. Theories of
symbol emergence in developmental psychology have been briefly mentioned
here. For further information, please see \cite{Callaghan2015}.}

\subsubsection{\addspan{Precursors of symbolic functioning}}
Bates et al.~\cite{bates1979} discuss that the onset and development of communicative intentions and conventional signals (such as pointing or crying for an unreachable toy), observed between 9 and 13 months, can be viewed as a precursor of symbolic communication. 

Before 9 months, infants can use some signals (such as crying). However, these signals are geared more toward the referent object than toward the adult that can help. By 9 months, signals and gestures become clear, consistent and intentional. For example, previously object-targeted cry signals are now aimed at adults. This can be verified through frequent eye contact and checks for feedback. Infants also start using clear and consistent vocal gestures, word-like sounds in request sequences, albeit in very context-bounded situations. Whereas these intentional communications and signals become more stable, regular, and predictable, Bates discusses that this is not considered to be standard symbolic communication: ``Conventional communication is not symbolic communication until we can infer that the child has objectified the vehicle-referent relationship to some extent, realizing that the vehicle, i.e., sign, can be substituted for its referent for certain purposes, at the same time the symbol is not the same thing as its referent''~\cite[p.~38]{bates1979}. For example, temporal separation of vocal gestures from the actual timing of the activity might be viewed as differentiation of the symbol from its referent, and considered to be truly symbolic activity, following Werner and Kaplan~\cite{werner1963}.

The period before the first words also corresponds to Piaget's sensorimotor stage~V, the stage where infants can differentiate means from ends and use novel means for familiar ends. After approximately 9 months of age, infants start using learned affordances to achieve certain goals, predicting desired changes in the environment to achieve the goals, and executing the corresponding actions~\cite{Piaget1952,Tomasello1999TheCultural,Willatts1999}. By 12 months, they can make multistep plans using learned affordances and perform sequencing of the learned action--effect mappings for different tasks. For example, they can reach a distant toy resting on a towel by first pulling the towel or retrieve an object on a support after removing an obstructing barrier~\cite{Willatts1984}. These data suggest the importance of early sensorimotor and social skill development in infants that are probably integrated into the language and reasoning network, where symbols and symbol manipulations play an important role. Learning a symbol system is also a continuous process, in which infants benefit from different principles such as similarity, conventionality, segmentation, and embodiment~\cite{Cangelosi2015}. 

\addspan{When we are away from ASS and do not assume pre-existing linguistic knowledge, our view has common grounds with Tomasello's usage-based theory of language acquisition~\cite{tomasello2009constructing,tomasello2009usage}. The usage-based approach to linguistic communication has two aphorisms, i.e., meaning is use, and structure emerges from use.
The first aphorism argues that the meanings should be rooted in how people use linguistic conventions to achieve social ends. The second argues that the meaning-based grammatical constructions should emerge from individual acts of language use. It is argued that, at around one year, children become equipped with two sets of cognitive skills, i.e., intention-reading and pattern-finding. 
Intention-reading performs the central role in the social-pragmatic approach to language acquisition. Pattern-finding includes categorization, analogy, and distributional analysis. This performs the central role in grammar construction. 
The usage-based theory of language acquisition is totally in line with the symbol emergence in cognitive systems, because both put importance on the bottom-up formation of symbolic/linguistic knowledge in cognitive systems. }

The first words, which can be seen as the clear indication of symbolic knowledge, are observed at around 12 months of age, when infants discover that things have names. For a thorough review of symbol emergence in infants and the role of nonlinguistic developments in symbol and language acquisition, see~\cite{bates1979}.

\subsubsection{\addspan{Language, graphical, and play symbolic systems}}

\addspan{The emergence of symbolic functioning is observed in infants in different domains, such as comprehension and production of verbal or sign language, gestures, graphic symbols, and (pretend) play. Efforts to clarify these range from investigating the capabilities of different age groups in
understanding the difference between symbol and referent using symbolic objects
such as pictures, maps, videos, and scale models \cite{Deloache2004} to assessing the abilities of
children in pretend play, which involves the ability to play with an object as if it
were a different one \cite{Tomasello1999}. Whereas the former is related to dual representation problem referring to understanding the symbol-referent relationship, the latter is related to the triune representation problem referring to using objects in a
symbolic way. It is important to note that related abilities of infants in different domains appear to reach common milestones. Gesture and language abilities, for example, follow the same stages, such as repetitive action, symbolic use, and combinatorial use on the same schedule \cite{BatesDick2002}. Both language and play acquisition begin with presymbolic structures, where action and meaning are fused (e.g., plays only related to infant's own sensorimotor repertoire), then abstract and agent-independent symbols emerge in both domains, and finally combinatorial use of the abstract symbols is observed \cite{Callaghan2002}.}

\addspan{Acquisition of language, among these domains, is special in humans. Young infants are capable of producing language-like sounds early in their first year: they can produce canonical babbles followed by protowords at month 7 and first words shortly after the first birthdate \cite{Adamson1995}. As we discussed in the previous subsection, the emergence of language comprehension is estimated to be around 9 to 10 months of age and language production in the first half of the second year \cite{Adamson1995}. The symbolic status of some of the early words might be specific to events and contexts, whereas other words might be used across a range of contexts \cite{Barrett1989}. With the development of naming insight early in the second year, when the child discovers that things have names, and names and their references are arbitrarily related, one can argue that symbolic functioning is in operation in the language domain.}

\addspan{Graphical or pictorial symbolic capability, however, is observed later: 2.5 year old children (but not younger ones) can retrieve hidden objects when the pictures of these objects are presented \cite{DeLoache1991}, or can match pictorial symbols with imaginary outcomes of actions given pictures of objects to which actions are applied \cite{Harris1997}.}

\addspan{Whereas there is no consensus on the existence of a single symbolic function/system that develops during early childhood and is realized in different domains, the symbolic systems in different domains (1) heavily influence each other, and (2) are significantly affected by scaffolding from the caregivers. In various experiments, it was shown that privileged and early developing symbol system of language is used by children in seemingly symbolic functions in the graphics and play domains. The object retrieval task based on pictorial symbols, for example, can be achieved by 2.5 years old children if the pictorial symbols can be distinguished verbally by the children (e.g., dog vs. cat - golden retriever vs. German shepherd) \cite{Callaghan2000}. Only after 3 years of age, children gain the capability to retrieve objects only using on their pictorial symbolic description without using language support. However, Callaghan and Rankin also showed that with more training, children could acquire the capability of pure pictorial symbolic functioning earlier than 3 years old \cite{Callaghan2002}.}

\addspan{From this discussion, we can conclude that early acquisition of the language symbol system compared to other domains is possibly due to the heavy parental scaffolding from birth \cite{Tomasello2009}, the symbol systems observable in different domains have similar developmental stages, they interact with each other, and finally their development depends on the amount of scaffolding and maturity of the system.}

\section{Problem history 2: Symbol emergence in artificial systems}
\label{sec:ProblemHistory2}

The main challenge concerning symbol emergence in cognitive developmental systems is to create artificial systems, e.g., robots, that can form and manipulate rich representations of categories and concepts. Our overview will concentrate on this problem in AI and machine learning.

\subsection{Artificial intelligence viewpoint: from tokens to symbols}\label{sec:hist_ai}

Originally, the idea of a symbol system in AI had its roots in mathematical/symbolic logic because all programming languages are based on this. Alas, this led to quite an inflexible image concerning the term {\it symbol}.  

Inspired by the work of Newell~\cite{Newell1980}, AI has tended to consider a symbol as the minimum element of intelligence and this way of thinking is hugely influenced by the physical symbol system hypothesis~\cite[p.116]{Newell1976} (see Section~\ref{sec:hist_cog} ):
\begin{quote}
``A physical symbol system consists of a set of entities, called {\it symbols}, which are physical patterns that can occur as components of another type of entity called an expression (or symbol structure). Thus, a symbol structure is composed of many instances (or tokens) of symbols related in some physical way (such as: token being next to another). At any instant of time, the system will contain a collection of these symbol structures.''
\end{quote}

This notion creates a strong link between a token in a Turing machine and a symbol. However, there are two major problems with this definition, generated by two strong underlying assumptions.

First, the physical symbol system hypothesis assumes that a symbol exists without any explicit connection to real-world information. Owing to the missing link between such symbols and the real world, such a physical symbol system is ungrounded, and therefore unable to function appropriately in complex environments. Later studies on symbol grounding~\cite{Harnad1990,taddeo2005solving}, symbol anchoring~\cite{coradeschi:2003:ras}, and the ``intelligence without representation'' argument~\cite{Brooks1991,clark94}, challenged the conventional understanding about the implementation of a symbol system.

Second, the token--symbol analogy is based on the assumption that a symbol deterministically represents a concept that does not change its meaning depending on the context or interpretation. However, this view is very different from that discussed in semiotics (see Section~\ref{sec:hist_semi}). There, the human (interpretant) takes a central role in the process of symbolization.

This generates confusion among researchers, especially in interdisciplinary fields such as cognitive systems, in which aspects of robotics and AI are combined with psychology, neuroscience, and social science, attempting to advance our understanding of human cognition and interaction by using a constructive approach. Programming-language-like symbols and human-meaning-like symbols are often conflated, and the term symbol is used to describe both (i) an atomic element of reasoning and (ii) a sign used in human interaction and communication.

Indeed, a general and practical definition of {\it symbol} that includes both aspects would be desirable. At the same time, the different meanings of this broad concept should be kept clear. In some of the literature on symbol grounding~\cite{coradeschi2013short} these two aspects are referred to as physical symbol grounding~\cite{vogt02psg} and social symbol grounding~\cite{cangelosi06ssg}. Another definition is proposed by Steels~\cite{Steels2008}, who refers to these different types of symbols as c-symbols and m-symbols, respectively, pointing out the ill-posed characteristics of the original symbol grounding problem.

\addspan{Studies in developmental robotics have been conducted to solve the symbol grounding problem. For example, Steels et al. have been tackling this problem using language games from the evolutionary viewpoint~\cite{steels2002grounding,steels2015talking}. Cangelosi et al. developed robotic models for symbol grounding~\cite{Cangelosi2000,Cangelosi2006}. For the further information please see~\cite{Cangelosi2015}.
Steels even said ``the symbol grounding problem has been solved~\cite{Steels2008}.'' However, it is still difficult for us to develop a robot that can learn language in a bottom-up manner and start communicating with people.}

To deal with these problems, we need to move on from the symbol grounding problem to symbol emergence by taking the process of bottom-up organization of symbol systems into consideration. This means that we need to develop a learnable computational model representing PSS that is also affected by social interactions, i.e., both physical \textit{and} social symbol grounding should be taken into consideration.

\subsection{Pattern recognition viewpoint: from labels to symbols}\label{sec:hist_pat}
Pattern recognition methods, e.g., image and speech recognition, have made great progress with deep learning for about five years~\cite{Krizhevsky2012,Dahl2012,DeepSpeech,goodfellow2016deep}.
Most pattern recognition systems are based on {\it supervised learning}. A supervised learning system is trained with input data and supervisory signals, i.e., desired output data. The supervisory signals are often called {\it label data} because they represent ground truth class labels.

What are symbols in pattern recognition tasks? Indeed, the class labels mentioned above were regarded as signs \addspan{representing concepts of objects} and the input data as \addspan{signs observed from} objects. Image recognition can, thus, be considered as a mapping between an object and a sign \addspan{representing a concept}. 
\addspan{Note that, here, they assume that there is a fixed concept defining the relationship between signs and objects. Additionally, the relationship is defined by human labelers from outside of the cognitive system, i.e., machine learning system. This means the pattern recognition problem does not treat the emergence of symbols.}
For example, when an image recognition system tries to recognize an apple in the pixels of an image, the system will try to map these pixels, i.e., signs, obtained from the object to the sign ``apple.'' The same holds true for speech recognition, in which sound waves are mapped to words.

\begin{figure}
\begin{center}
\includegraphics[width=80mm]{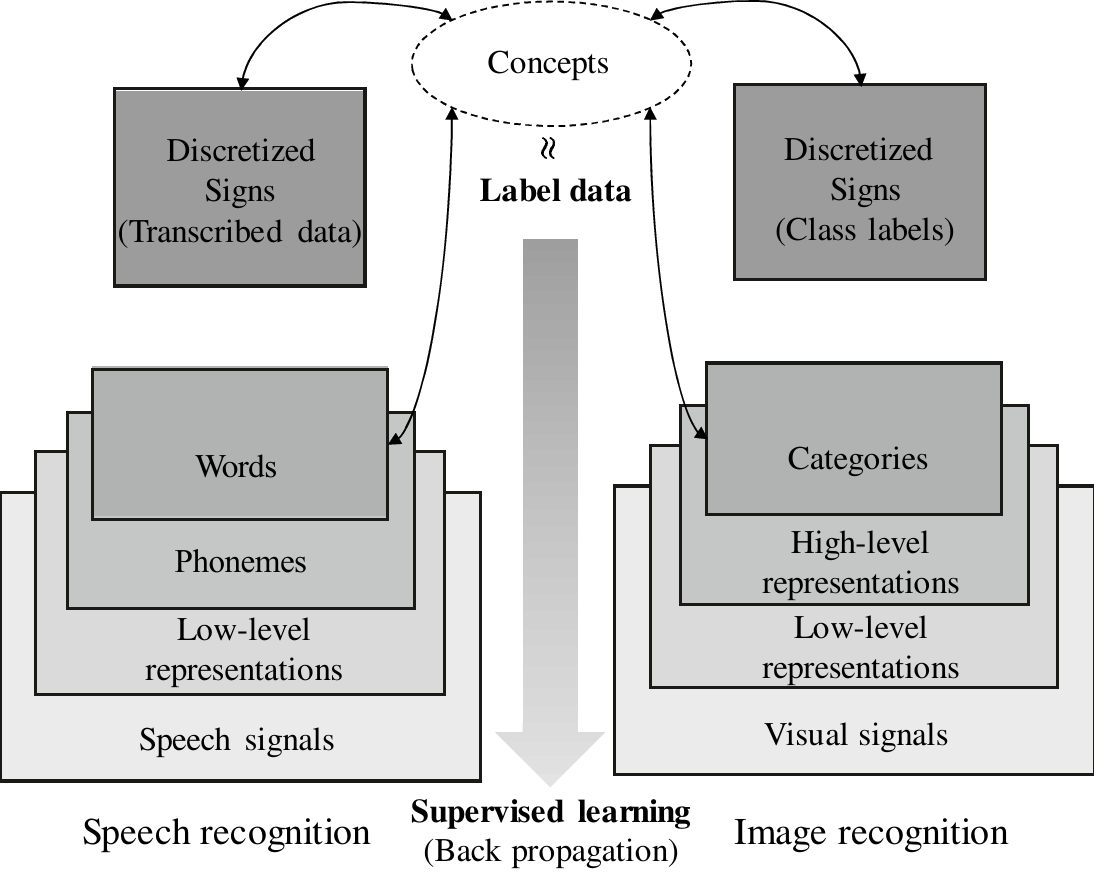}
\caption{Pattern recognition systems, e.g., image recognition and speech recognition systems, usually presume a hierarchical structure of the target data. Label data, usually corresponding to a sign representing a concept, is given by supervision in the training process.}
\label{fig:vision_speech}
\end{center}
\end{figure}

The general pipeline for image or speech recognition is shown in Fig.~\ref{fig:vision_speech}. Conventional recognition systems use pre-defined low-level features, mostly defined by a human, to structure the input data prior to recognition. Modern deep learning performs \emph{end-to-end} learning and features exist only implicitly therein (as the activation structures of the hidden layers)~\cite{Krizhevsky2012}, but it also makes supervised-learning-based pattern recognition a well-defined problem. Alas, this assumption led to large side effects on our way toward understanding symbol systems. From the above, it is clear that this type of pattern recognition essentially assumes static symbol systems.

\addspan{However, when we look at the interior of the pattern recognition systems using deep neural networks, we can find that neural networks form internal representations dynamically, even though the learning process is governed by static symbol systems.}
For example, it was found that convolutional neural networks (CNNs) form low-level feature representations in the shallow layers and high-level feature representations in the deep layers~\cite{Krizhevsky2012}.
\addspan{Those features themselves are not symbols. However, to realize symbol emergence in cognitive systems the cognitive dynamics that form rich internal representations are crucial.}  

In summary, pattern recognition has been providing many useful technologies for grounding labels (signs) by real-world sensory information. However, these methods generally assume a very simple dyadic notion about symbols, i.e., the mapping between objects and their labels. From a developmental viewpoint, a neonatal development system cannot get label data directly as discretized signals input to its internal cognitive system. Thus, most supervised-leaning-based pattern recognition system cannot be regarded as a constructive model of human development \addspan{and symbol emergence}.

\subsection{Unsupervised learning viewpoint: from multimodal categorization to symbols }\label{sec:hist_uns}

It has been shown by a wide variety of studies that it is possible to create \addspan{an internal representation system that internalizes a symbol system, reproduces categories, and preserves concepts} by using unsupervised clustering techniques~\cite{Bishop,Taniguchi2016SER}. One example where the complete chain from sensorimotor experience to language symbols has been addressed is the work of Nakamura et al.\, as discussed in the following.

The central goal of the studies was that a robot creates a certain conceptual structure by categorizing its own sensorimotor experiences. When this happens using \textit{multimodal} sensorimotor information, this can be regarded as a PSS model. In addition, the use of multimodal sensorimotor information for forming categories is important because it enables the cognitive system to perform cross-modal prediction. For example, humans can usually predict whether it is hard or soft simply by looking at a particular cup, e.g., paper or porcelain. This means that you can predict tactile information only from visual information through such a cross-modal inference. \addspan{This type of embodied simulation is considered as a crucial function of PSS.}

This has been called \emph{multimodal object categorization}, realized by applying a hierarchical Bayesian model to robotics~\cite{Nakamura2007}. It can be achieved by multimodal latent Dirichlet allocation (MLDA), which is a multimodal extension of latent Dirichlet allocation (LDA)---a widely known probabilistic topic model~\cite{Blei2003}. This way, a robot can form object categories using visual, auditory, and tactile information acquired by itself. 

An important point here is that the sensorimotor information acquired by the robot generates categories that are, at first, only suitable for \textit{this} robot. However, it is also possible to add words as an additional mode of information~\cite{Nakamura2007}. This way, the created categories become much closer to human categories, owing to the co-occurrence of words and the similar multimodal sensory information. Thus, this represents a probabilistic association of words, signs, with a multimodal category, which related to a certain concept---a process of symbol formation. This also makes it possible to formulate a semantic understanding of words as a prediction of multimodal information through categories.

One interesting extension of this model concerns the self-acquisition of language combining automatic speech recognition with the MLDA system. Here, unsupervised morphological analysis~\cite{Nakamura2014} is performed on phoneme recognition results in order to acquire a vocabulary. 
The point of this model is that multimodal categories are used for learning lexical information and vice versa. This way, the robot could acquire approximately 70 appropriate words through interaction with a user over 1~month~\cite{Aoki2016}. This suggests that unsupervised learning, e.g., multimodal object categorization, can provide a basic internal representation system for primal language learning.

There have been many related studies. For example, Taniguchi introduced the idea of \emph{spatial concept} and built a machine learning system to enable a robot to form place categories and learn place names~\cite{Akira2016,Akira2017}. Mangin proposed multimodal concept formation using a non-negative matrix factorization method~\cite{Mangin2015}. Ugur et al.\ studied the discovery of \textit{predictable} effect categories by grouping the interactions that produce similar effects~\cite{Ugur2012IROS}. They utilized a standard clustering method with varying maximal numbers of clusters, and accepted a number of clusters only if the corresponding effect categories could be reliably predicted with classifiers that take object features and actions as inputs. \addspan{Predictability also performs a central role in the series of studies by Tani et al.~\cite{tani2016exploring}.}

All of these studies suggest that bottom-up concept formation, using sensorimotor information obtained by a robot itself, is possible, and this appears to be a promising avenue for future research.

\subsection{Reinforcement learning viewpoint: from states and actions to symbols} \label{sec:hist_rl}
Whereas the aforementioned work focuses on finding structure in sensory information, decision-making and behavior learning are fundamental components of cognitive systems as well. This problem is often formulated as a reinforcement learning problem~\cite{sutton1998}. 
The goal of a reinforcement learning agent is to find an optimal policy that can maximize the total accumulated reward obtained from its environment. \addspan{In the context of reinforcement learning, the term {\it symbol} has been used related to state and action abstraction with the belief that symbol is an abstract discrete token under the conceptual influence of symbolic AI.}

Reinforcement learning is most commonly formulated as learning over a Markov decision process, which often assumes that an agent has a discrete state--action space. However, if its action and state spaces are too large, the curse of dimensionality prevents the agent from learning adequate policies. Therefore, how to design a compact and discrete state--action space for an agent, or how to enable the agent to perform state and action abstraction were regarded as crucial problems in the history of reinforcement learning. Discrete states and actions formed through some sort of abstraction, especially when they are meaningful/interpretable units of states or actions, have often been regarded as symbols or concepts in a naive way. \addspan{However, the internal feature representations, i.e., abstracted states, can not have any nature of symbols if they do not have semiosis, given our definition of symbols in this paper. }

State abstraction was commonly treated as a discretization of the state space described by a discrete set of symbols, as in some form of tile coding~\cite{sutton1996}. Another way of abstraction is to use a function approximation that can map continuous states or a large discrete state space directly to action values. For example, these abstract states may be represented by activations in neural networks, i.e., distributed representations~\cite{tesauro1995}. This has more recently led to an explosion of work in deep reinforcement learning (DRL), which has involved behavior learning against discovered, higher-level representations of sensory information~\cite{mnih2015human}. In the context of DRL, state abstraction is represented in the same way as high-level feature representation in CNNs in pattern recognition systems~\cite{Krizhevsky2012}.

Action abstraction has similarly provided a useful mechanism for learning. Operating and learning exclusively at the level of action primitives is a slow process, particularly when the same sequence of actions is required in multiple locations. Action hierarchies can then be defined through \emph{options} or macro-actions~\cite{sutton1999}, which are defined as local controllers and constructed from primitive actions (or simpler options). These can then be treated as new actions, and can be used in planning as single atoms or symbols referencing more complex behaviors~\cite{barto2003}. In the reinforcement learning literature, these are learned as behaviors with some functional invariance, for example as controllers that lead to bottleneck states~\cite{stolle2002}, or as commonly occurring sub-behaviors~\cite{ranchod2015}. Additionally, in more recent work, hierarchies of actions can be defined to allow for concurrent, rather than sequential, execution~\cite{saxe2017}.

These different forms of abstraction are not only useful for improving the tractability of problems, but they also provide a useful form of knowledge transfer, where solving one task provides an encoding of the environment that facilitates better performance on the next task~\cite{taylor2009}.
Mugan and Kuipers~\cite{Mugan2012} implemented a system that learns qualitative representations of states and predictive models in a bottom-up manner by discretizing the continuous variables of the environment.
In another study, Konidaris et al.\ studied the construction of ``symbols'' that can be directly used as preconditions and effects of actions for the generation of deterministic~\cite{Konidaris2014} and probabilistic~\cite{Konidaris2015} plans in simulated environments. \addspan{Note that the usage of the term {\it symbols} in this study simply corresponds to internal feature representations, even though it looks like ``symbols'' in terms of symbolic AI.}

These studies all investigated how to form \emph{symbols}, i.e., internal feature representations, in the continuous sensorimotor space of the robot. However, complex \emph{symbols} can be formed by combining predefined or already-learned symbols. 
For example, Pasula et al.~\cite{Pasula2007} and Lang et al.~\cite{Lang2010} studied the learning of symbolic operators using predefined predicates.
Ugur et al.\ re-used the previously discovered symbols in generating plans in novel settings~\cite{Ugur2015,Ugur2015Humanoids}.

In the reinforcement learning literature, most work related to symbol emergence was about the formation of internal feature representation systems, in particular state--action abstractions, for efficient behavior learning. \addspan{State and action abstraction is regarded as a part of symbol emergence, but not symbol emergence itself.} 

\addspan{Recently, several studies extended the framework of reinforcement learning and enabled an agent to learn interpretation of linguistic, i.e., symbolic, input in the context of DRL~\cite{hermann2017grounded}. Most of the studies remain at the preliminary stage from the viewpoint of natural language understanding. However, this is also a promising approach to model symbol emergence in cognitive developmental systems.} 

\subsection{Dynamical systems viewpoint: from attractors to symbols}\label{sec:hist_dyn}
Any agent (human, animal, or robot), which is physically embodied and embedded in its environment, can be described as a continuous dynamical system. The question, which has been addressed by several researchers, is whether discrete states or \textit{proto-symbols} can emerge and be identified in such systems.

The notion of \textit{attractors} in a nonlinear dynamical system provides a natural connection: the attractor (no matter whether a fixed point or a limit cycle) can be seen as a discrete entity and there would typically be only a limited number of them. Pfeifer and Bongard~\cite[pp.~153--159]{PfeiferBongard2007} offer the example of a running animal or robot, where the different gaits (such as walk, trot, gallop) would correspond to attractors of the brain-body-environment system. This could be identified by an outside observer, but, importantly, also by the agent itself in the sensorimotor space accessible to its neural system. Furthermore, one can imagine other, downstream networks in the agent's brain that would operate with these proto-symbols, perhaps instantiating a primitive form of ``symbol processing.'' The gaits and the sensorimotor structure they induce would influence the dynamics of the downstream networks, and conversely the low-level network could be influenced by the ``symbolic'' level through the motor signals, corresponding to gait selection on the part of the agent.
A related but more formal account of ``dynamics-based information processing'' dealing with mapping between motion space and symbol space in a humanoid robot was put forth by Okada and Nakamura~\cite{Okada2004_ICRA}.

This question is also addressed by Kuniyoshi et al.~\cite{Kuniyoshi2003}, who make use of the mathematical concept of structural stability: the claim is that a ``global information structure'' will emerge from the interaction of the body with the environment. Kuniyoshi concludes that because of the discrete and persistent nature, one can identify this global information structure with the notion of \textit{symbols}. This account is further extended to the interaction of multiple agents, moving from \emph{intra-dynamics} (like the gaits as attractors) to \emph{inter-dynamics}. 

Whereas these studies have addressed artificial agents, we note that their viewpoints are also strongly related to the above discussion about the mirror system in primates as a possible neural instantiation of a proto-symbol structure.

\section{Integrative viewpoint}\label{sec:IntegrativeView}

In this section, we integrate the viewpoints described in Sections~\ref{sec:ProblemHistory1} and \ref{sec:ProblemHistory2}, and provide a unified viewpoint on symbol emergence.

\subsection{Wrapping things up (from the perspective of PSS)}
It has been a generally accepted view in AI (Section~\ref{sec:ProblemHistory2}) and cognitive science (Section~\ref{sec:hist_cog}) to regard symbol systems as internal representation systems. Cognitive neuroscience also follows this way of thinking (\ref{sec:hist_neuro}). 
\addspan{However, this idea has been affected by the misconception that ``symbols are a discrete token in our mind'' given by the physical symbol system hypothesis. There are no discrete tokens in our brain, and a symbol in our society is not such a type of static thing as a broad range of social, cultural, and linguistic studies suggest.
The misconception even introduced confusion in the usage of related terms, e.g., \emph{internal representation system}, \emph{concept}, \emph{category}, \emph{symbol}, and \emph{feature representation}. They are used with different meanings by different people and scientific communities related to symbol systems.}
\delspan{ Many times a fundamental misunderstanding about symbol system lies in the fact that the terms embedded therein, \emph{internal representation}, \emph{concept}, \emph{category}, \emph{symbol}, etc., are used with different meanings by different people (and by different scientific communities).}

Furthermore, we discussed that there are two types of usages of the term, symbol system, considered by the different fields: symbol systems in society as compared to the internal representational systems in our brain. Whereas both are clearly linked to each other through processes of externalization, e.g., language generation and understanding, they are also often confused in the literature. In particular, symbolic AI and in turn, cognitive science, have been totally confusing them.

In this paper, we start with PSS. The focus on PSS chosen here is motivated by its more dynamic nature that better fits the requirements of cognitive developmental systems.
Concerning PSS, we can summarize that this theory assumes that internal representation systems, i.e., PSSs, are formed in a bottom-up manner from perceptual states that arise in sensorimotor systems. A perceptual state is formed by a neural representation of the physical input and this can also be supplemented by a conscious experience, a subset of perceptual states is stored in long-term memory. Thus, these perceptual memory contents can function as concepts, standing for referents in the world, constituting the representations that underlie cognition. 

The bottom-up development assumed by PSS leads to the fact that such a PSS can be regarded as a self-organization process of multimodal sensorimotor information in our brain. This view is quite similar to the dynamics of the schema model, proposed by Piaget to explain human development during the \emph{sensorimotor period}. 

\delspan{PSS considers more adaptive and dynamic properties of symbol systems in our cognitive system than amodal symbol systems (ASS).  It remains, however, unclear to what degree this would cover the full semiotic dynamics assumed by Peirce, i.e., semiotics. Models need to be created to address this issue.}

What we need here for an integration of all this is as follows. (1) If we distinguish symbol systems in society and those in our mental systems, and (2) if we consider the dynamics and uncertainty in symbol systems, and (3) if we also take the context dependency of symbols (i.e., semiosis) into account, and, finally, (4) if we succeed in combining the different (society versus internal) levels of symbolization, then we will get closer to an integrative viewpoint of a symbol system.

\subsection{\addspan{Symbol emergence systems}}
\delspan{{\it Symbol emergence as a semiotic process}}
\begin{figure}
\begin{center}
\includegraphics[width=80mm]{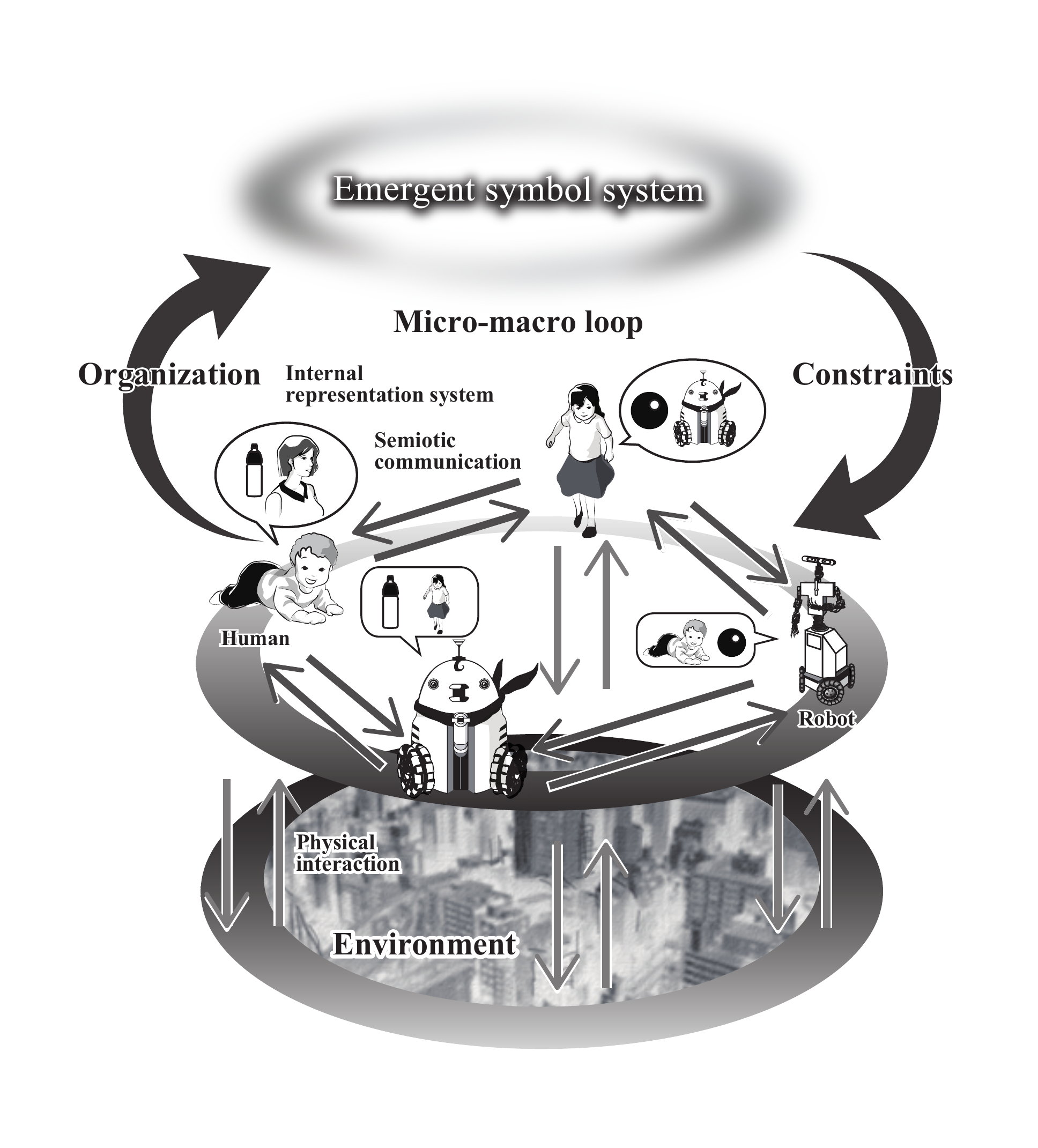}
\caption{Overview of a symbol emergence system~\cite{Taniguchi2016SER}. Cognitive developmental systems including autonomous robots need to be elements of the system to perform semiotic communication.}
\label{fig:ses}
\end{center}
\end{figure}

\begin{figure*}
\begin{center}
\includegraphics[width=160mm]{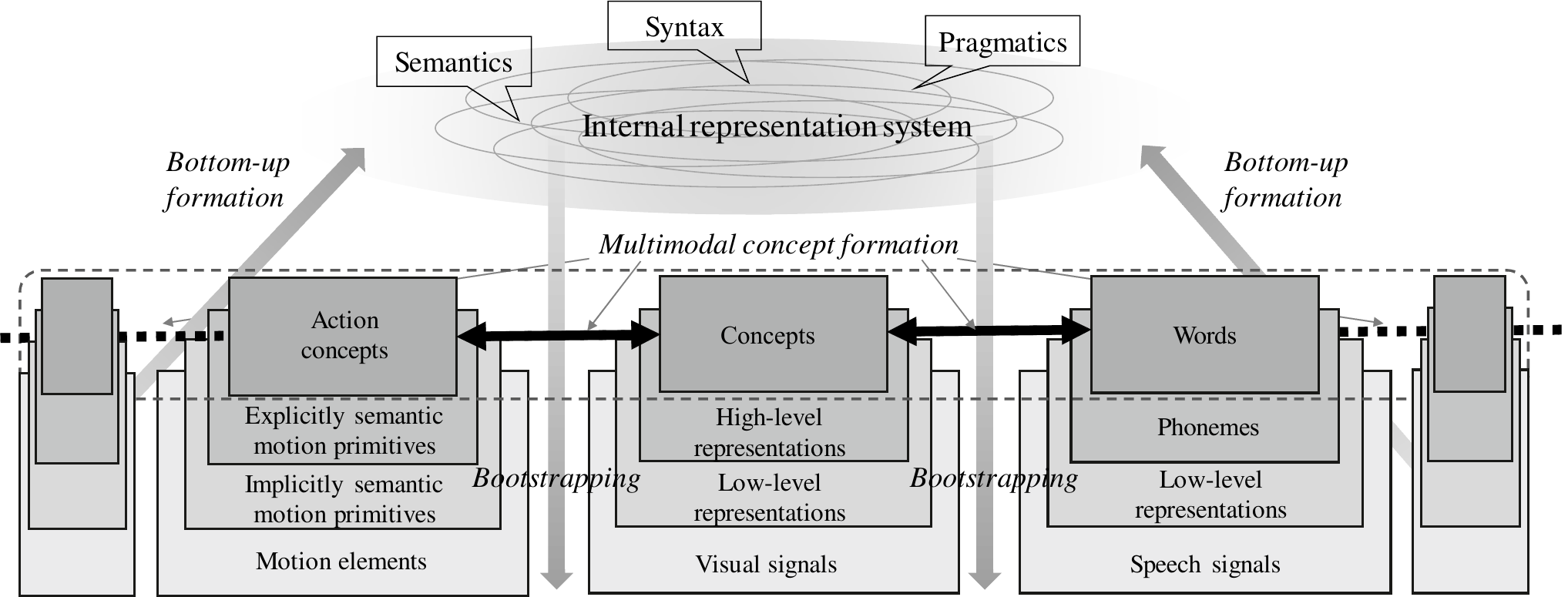}
\caption{Multimodal fusion of sensorimotor information and self-organization of concepts and symbols. }
\label{fig:multimodal_fusion}
\end{center}
\end{figure*}

Figure~\ref{fig:ses} shows a schematic illustration of symbol emergence systems, describing a potential path to symbol emergence in a multiagent system~\cite{Taniguchi2016SER}. In this figure, we distinguish between an internal representational system and a symbol system owned by a community or a society.

At the start, human infants form their object categories and motion primitives through physical interaction with their environment. Soon, however, they will also generate signs, e.g., speech signals and gestures, to accompany their actions, or to express their wishes. Alongside the development of the infants' ability to generate signs, their parents and other people learn how to interpret, but also correct, the signs of the infants. Thus, this type of mutual symbol negotiation process gradually realizes semiotic communication.

The child's symbols, that had originated from bottom-up sensorimotor processes, will in this way drift toward the symbol system shared by their community, at first the infant and parents. Hence, language learning can be regarded as \addspan{both the bottom-up formation process and} the internalization of a symbol system shared in a community. \addspan{This view aligns with Tomasello's usage-based theory of language acquisition~\cite{tomasello2009usage}.}  

In general, all such systems evolve within their communities and they follow certain rules, which impose constraints on all agents that make use of them. Without following these rules, e.g., semantics, syntax, and pragmatics, communication would be hard or even impossible. Any agent is free to form categories and signs solely based on perceptions and actions, but when an agent needs to communicate using signs, they must be standardized throughout its community.
\addspan{Therefore, both bottom-up organization and top-down constraints are crucial in our symbol systems. This leads us to the concept of {\it symbol emergence systems}.}

The concept of {\it emergence} comes from studies of complex systems. A complex system that shows emergence displays a macroscopic property at some higher level that arises in a bottom-up manner based on interactions among the system's lower-level entities. In turn, the macroscopic property constrains the behaviors of the lower-level entities. Ultimately, this so-called \textit{emergent} high-level property endows the system with a new function, which cannot be generated by the individual low-level components on their own. The interaction between high and low levels in such a system, consisting of bottom-up organization and the top-down constraining, is called a {\it micro--macro loop}.

Figure~\ref{fig:ses} depicts this for our case. The sensorimotor and also the direct social interaction processes of the individual reside at the lower levels. The resulting symbol system represents a higher-level structure, which---as described above---constrains the lower levels. Hence, such symbol systems may well be regarded as an emergent structure with a typical example of a micro--macro loop.
\addspan{Therefore, it may be better to call it an {\it emergent symbol system}, distinguishing it from conventional physical symbol system in the old-fashioned AI.
As a whole, the multi-agent system can be regarded as a complex system with emergent property producing functions, i.e., semiotic communications. Therefore, it is fair to say this is a type of an {\it emergent system}. Thus, we call this complex system a {\it symbol emergence system}.}   

\addspan{To become an element of a symbol emergence system, the element, e.g., a person or a cognitive robot, must have the cognitive capability to form an internal representation system, i.e., PSS, and to perform symbol negotiation.} 
Figure~\ref{fig:multimodal_fusion} shows an abstract diagram of a partial structure of a cognitive system required for an agent participating in a symbol emergence system. We already discussed above, when summarizing PSS, how individual cognitive traits, some of which are shown in this figure, arise based on multimodal sensorimotor information, and also how internal representations are formed as multimodal concepts. Hence, information from different modalities is highly intertwined. The figure also emphasizes several high-level aspects such as syntax, semantics, and pragmatics. For us, syntax not only plays a role in language but it has been suggested that syntactic structures are of fundamental importance for our action planning capabilities, for example in manipulation (see Section \ref{sec:hist_evo} and~\cite{pastra11grammar}). Conversely, it has also been hypothesized that the evolution of syntactic planning capabilities became a scaffold of the evolution of language.
\addspan{Some studies have already shown that structures in an internal representation system analogous to action grammar emerge as distributed representations in neuro-dynamic systems~\cite{tani2016exploring}.}
One should likely think of this as an upward-winding spiral, where one process drives the other and vice versa. \addspan{This knowledge should be self-organized as an internal representation system in a distributed manner. We expect that we can develop neural network-based and/or probabilistic models to artificially realize such learning dynamics in the near future.}

\subsection{Redefinition of the terminology}
In this section, we try to identify the differences between terms related to \emph{symbols}, from the viewpoint of a symbol emergence system, to address possible confusion in future interdisciplinary research. 

\subsubsection{Concept and category}
These two terms are---as mentioned above---often used interchangeably across the literature, even in cognitive science. Here, we would like to offer a practical solution to disentangle this.

\underline{Category:} 

A category corresponds to a referent, i.e. an object in semiosis, where exemplars of the object are invoked by a perception or a thought process. Categories are influenced
by contextual information. Categories can be formed and inferred by an agent from sensorimotor information, i.e., features, alone without any labeled data. For example, the MLDA-based approach can categorize objects into object categories probabilistically~\cite{Nakamura2009,MHDP,Ando,Araki2012,Nakamura2011,Nakamura2012b,Nakamura2015}. A multimodal autoencoder can also perform a similar task. 
In order to make sense for an agent that acts in some world, categories need to be based on real-world experiences. This can either happen directly through forming a category from sensorimotor inputs, or indirectly through inferring a category from entities in memory.

Categories can be memorized in long-term memory, but  
importantly, they are not static. Instead they update
themselves in long-term memory by the interactions between actual
objects, instances, relations, or events in the world. We have
many relatively static categories in our memory system
and society (such as \emph{fruit}, \emph{car}, etc.), where the properties that define these
categories are more or less static; these help us structure our
world. 
We can also dynamically infer or generate categories in a real-time manner. Ad hoc categories, e.g. things to take on a camping trip, are a good example for this.
Furthermore, we assume we can distinguish items belonging to one category from those in another one using categorical recognition. 

In summary, categories are very much an exemplar-based idea, and always have an inference process using category-determining features that make an object a member of the category. This process can be regarded as semiosis. 

\underline{Concept}: 
\addspan{A concept, on the contrary, is usually regarded as a context-independent internal representation embedded in long-term memory, as opposed to a category which is an exemplar-based idea.
When we follow the idea of PSS, cross/multimodal simulation is a crucial function of a concept, i.e., perceptual symbol. Here, the concepts embedded in long-term memory allow the cognitive system to predict other modality information from the given modality information through inference. When we hear a speech signal  ``apple'', we can infer its color, shape and taste via a concept formed using multimodal sensorimotor information. On the contrary, we can infer the object's name, i.e., ``apple'', from its color, shape and taste information.
The semantic essence of a category can be used to infer categories and other useful information.
This definition fits with the idea of PSS.} 

Difference between semantic essence of exemplars, and categories can also be understood by means of an example.
Let us consider the category of all possible Go games, determined by their shared features. As we
know, AlphaGo~\cite{Silver2016} can by now beat the best Go players in
the world. Within AlphaGo the representation of how to play a game is encoded in the network weights. 
Neither AlphaGo itself nor anyone else can extract from these weights and explain the goal of Go and the game strategiesr required to achieve it. 
Goal and strategy constitute in this example what we would call the essence of the Go-game category. We would argue, to be able to
say that an agent has a oncept of Go, those strategies need
to be represented in an extractable and explainable way. The fact that we,
humans, can talk about strategies of playing Go makes clear
that for us that this is true. We, after learning the game, can
extract strategies of play. It is, however, not the fact that we
do extract them explicitly, but rather the fact that they can
be extracted \addspan{or explained linguistically. This means concepts should have connections with explanations using language, i.e., symbol systems.}
Furthermore, this agent may also be able to do the same for Checkers, Monopoly, Chess, and some other board games.
This agent would be able, for example by clustering in the space of game strategies, to distinguish these different games.
In consequence, it can form the Category of (those) Board Games. We ask whether such an agent also have arrived at
the concept of Board Games (assuming such a concept exists)? The answer should in the first place be ``no''. At this stage, the
agent merely possesses a collection of exemplars representing the category.
To form a Concept, the agent would need to be able to extract the fact which strategies are an essential ingredient of board games and are required to achieve the goal of the game. Moreover, it would need to be able to represent other essential characteristics of board games such as that they are fundamentally 2D, the surface is always divided into some segments, the pieces are moved horizontally, etc., even if these were not recruited by the category-forming clustering procedure.

\addspan{In summary, the concept refers to an internal representation which goes beyond the feature-based representation of a category. To form a concept of a category an agent needs to be able to extract the semantic essence of the category, and to connect category and semiosis. 
At the same time, the concept conversely needs to be able to infer categories from signs or linguistic explanations.}
Concepts can be used to form yet another, higher-level category, where concept formation could then set in again.
\addspan{When we develop a cognitive system that can form concepts, we can use not only probabilistic models, but also neural networks. For example, a relational network can infer grounded meaning in sentences by extracting some conceptual structures without using discrete symbol structures in their internal representations~\cite{raposo2017discovering}.} 

We believe that the distinction offered here between category
and concept captures many of the aspects of these two entities discussed above. These definitions can be phrased
in computer (or network) terms which should be useful for our purpose of helping to develop improved cognitive developmental systems.

\subsubsection{\addspan{Feature representations}}\delspan{{\it Feature hierarchy:}}
\addspan{For performing recognition and other tasks, a cognitive system needs to perform a hierarchical feature extraction and obtain various feature representations.}
Raw data, e.g., sound waves and images, do not represent a category or a concept on their own. Let us consider image classification as an example. From raw data, one can first extract local low-level features (e.g.,\ edges, textures, etc.). Modern CNNs can automatically generate low-level features represented by the activations of their lower layers, which, however, are usually not human-understandable. Higher layers of a CNN will combine many low-level features of the image, creating higher-level features. A CNN thereby creates a feature hierarchy, and it has been demonstrated~\cite{Krizhevsky2012}  
that some high-level features correspond to meaningful entities, e.g., a human face or a human body. Thus, high-level features represent more complex information than low-level features or raw signals. However, it goes too far to assume that such high-level features are categories or concepts, because so far all of the above processes are strictly supervised and not grounded in the perception of a cognitive agent. In addition, there is no semiotic process present when such a feature hierarchy is formed. Still, feature hierarchies are, as a result of a data structuring process, a very useful front-end for cognitions.
\addspan{Many studies on ``symbolization'' in previous AI-related studies can be regarded as high-level (discrete) feature representation learning. It is too naive to refer to simple representation learning as symbol emergence.}

\subsubsection{\addspan{Internal representation system}}
\delspan{{\it Internal representation/symbol system: }}
The term ``internal representation'' has a very broad meaning. 
In a cognitive model, any activation pattern of a neural network and any posterior probability distribution in a probabilistic model can be regarded as an internal representation in a broad sense. \addspan{Note that concepts encoded in our brain system can be also regarded as internal representations as well.}

However, from the viewpoint of this review, we would posit that the term \textit{internal representation system} refers to a structured system that is formed \addspan{by interacting with symbol systems in the society and somehow internalizing it. This 
retains categories and concepts. It also retains knowledge of syntax, semantics and pragmatics.}

This is organized by multimodal perceptual information obtained from its sensorimotor input and from the interaction with the society in which the agent is embedded. As discussed above, these bottom-up principles of organization let an internal representation system \textit{emerge}. Such an internal representation system \addspan{can be regarded as a PSS}. \addspan{Note that the internal representation systems are never amodal symbol systems.}

\subsubsection{\addspan{Emergent symbol system}}
\addspan{When we open a dictionary, we can find many words representing concepts and categories and their relationships. The meaning of each word is described by some sentences, i.e., syntactically composed words. In classical AI, they attempted to give a type of knowledge to robots. That led to amodal symbol systems. We agree that this type of ``symbol system'' does exist in society, but disagree that we have such physical tokens in our cognitive system.}

\addspan{Through negotiation between cognitive agents, the standard meanings of words and sentences are gradually organized in a society or a community. Note that the symbol system itself dynamically changes over time in a small community and in the long term. However, it looks fixed in a larger community and over the short term. Therefore, people tend to misunderstand there is a rigid and fixed symbol system. }

\addspan{We argue that the symbol system in our society has an emergent property. The symbol system is formed through interaction between cognitive systems that have capability of bottom-up formation of internal representation systems. When the cognitive systems try to communicate with others, they have to follow the symbol system. The symbol system gives constraints to the agents' symbolic communications. The symbol system drifts over time. We call this structural abstract entity an {\it emergent symbol system} to emphasize its emergent property. Essentially, all symbol systems must be emergent symbol systems.}

\subsubsection{Other related terms}
In this section, we mention several terms that also tend to be confused with symbols or symbol systems, although they are clearly different. 

\underline{Discrete state}: As discussed above, symbols have strong connections with categories and concepts. These are by construction somehow discrete in the sense that every category/concept has a boundary to other categories or concepts. This definition goes beyond the token-like, more static discreteness assumed by amodal symbol theories. The PSS perspective, as adopted here, allows boundaries to move and be vague. 
\addspan{This tends to give people the misbelief that a discretized state is a concept or a symbol, which is totally wrong. Additionally, there are many studies that call discretization symbolization. 
Symbolization may introduce some discrete structure into the world because of its nature of categorization. However, discretization alone cannot be regarded as symbolization.}

\underline{Word}: 
A word is the smallest element that is written or uttered in isolation with pragmatic and semantic content in linguistics. It also becomes a syntactic element in a sentence, and it consists of a sequence of phonemes or letters.

In an ASS (or even in formal logic), we tend to misunderstand this term to the degree that we would naively think that a word, e.g., ``apple,''  itself conveys its own meaning. However, \addspan{as studies following the symbol grounding problem suggested}, to interpret the meaning of a word, the system needs to ground the word by relating it, possibly via category and/or concept, to its sensorimotor experience. Alternatively, the system could also borrow meanings from other grounded words using syntactic rules. These processes are called \emph{sensorimotor toil} and \emph{symbolic theft}, respectively~\cite{Cangelosi2000}.

\addspan{Moreover, the arbitrariness of labels and concepts suggested in semiotics should be considered. The word representing an apple, concept and category of the word ``apple'' can be different in different languages and regional communities. Note that a word is just an observed sign. It itself it is not a concept or a category. Of course, it is not a symbol.}
 
\underline{Language}:
Human beings are the only animals that can use complex language, even though robots might be able to use some type of reduced language in the near future. 
Generally, symbol systems do not need to have all properties of a language. Language involves syntax, semantics, pragmatics, phonology, and morphology. Conventional formal logic has regarded language as a formal system of signs governed by syntactic rules for combining symbols, i.e., symbols in an ASS, by extracting some logical part of language. The syntactic characteristic is a crucial property of languages. However, other aspects of languages are also important to develop a robot that can communicate with people and collaborate in a natural environment. To do this, the robot should have cognitive capabilities allowing it to learn lexicons, semantics, syntax, and pragmatics through sensorimotor interaction with its environment and other robots and humans. This is one of the central challenges in cognitive developmental robotics.

\section{Challenges}\label{sec:Challenges}
\subsection{Computational models for symbol emergence and cognitive architecture}
\addspan{Developing a computational model that realizes symbol emergence in cognitive developmental systems is a crucial challenge. Many types of cognitive architectures, e.g., {ACT-R}, {SOAR}, and Clarion, have been proposed to give algorithmic descriptions of phenomena in cognitive science\cite{sun2016anatomy,taatgen2008constraints}. 
They described possible psychological process in interaction between symbolic and sensorimotor computations.
However, their adaptability and developmental nature are limited. We need a cognitive architecture that enables robots to automatically learn behaviors and language from a sensorimotor system to allow them to play the role of an agent in a symbol emergence system.}

\addspan{Tani et al. have been studying computational models for internal representation systems based on self-organization in neuro-dynamic systems~\cite{tani2016exploring,Tani2014}. Their pioneering studies have shown that robots can form internal representations through sensorimotor interaction with their environment and linguistic interaction with human instructors. The central idea is that continuous neural computation and predictive coding of sensorimotor information is crucial for internal representation formation.
Recently, we can find many studies that apply deep learning scheme to multimodal human-robot interaction including linguistic modality~\cite{heinrich2018interactive,pan2017video,lala2017detection,shridhar2017grounding,interact_picking18}.}
\addspan{Recent studies on artificial general intelligence based on DRL follow a similar idea. Hermann et al. showed that DRL architectures can integrate visual, linguistic, and locational information and perform appropriate actions by interpreting commands~\cite{hermann2017grounded}.}
\addspan{Taniguchi et al. have been using probabilistic generative models (i.e., Bayesian models) as computational models of cognitive systems for symbol emergence in robotics~\cite{Taniguchi2016SER}. It is occasionally argued that their models assume discrete symbol systems in a similar way as ASSs. However, this involves misunderstanding. A cognitive system designed by probabilistic generative models has its internal states as continuous probability distributions over discrete nodes in the same way as a neural network has its state as activation patterns over discrete nodes, i.e., neurons.}

\addspan{Neural networks can be trained using stochastic gradient descent and related methods, i.e., backpropagation, and this allows a system to have many types of network structure and activation functions. Recently, neural networks with external memory are gaining attention~\cite{graves2016hybrid,graves2014neural}. 
Therefore, if we have a large amount of data, a neural network can organize feature representations adaptively through supervised learning. This is an advantage of neural networks. However, in an unsupervised learning setting, it is still hard to introduce a specific structure in latent variables.}
\addspan{In contrast, probabilistic generative models can easily introduce structural assumptions of latent variables, and are suitable for unsupervised learning. Cognitive developmental systems should be essentially based on unsupervised learning, because internal representations should be self-organized through interactions in an unsupervised manner. They have been making use of probabilistic generative models and produced many unsupervised learning-based cognitive developmental systems. However, at the same time, the inference procedures of the models were not so flexible compared to neural networks, i.e., backpropagation. MCMC and variational inference, practically, tended to require the use of conjugate prior distributions, e.g., Gauss--Wishart, and Multinomial-Dirichlet distributions~\cite{Bishop}. This has been preventing their models from having feature representation learning capabilities.}

\addspan{However, it should be emphasized that the employment of probabilistic generative models or neural networks, e.g., neuro-dynamics systems by Tani, is not a binary choice. For example, Kingma et al. introduced autoencoding variational Bayes and gave a clear relationship between variational inference in probabilistic generative models and autoencoding using neural networks~\cite{kingma2013auto}. They proposed variational autoencoder (VAE) as an example of this idea. By integrating VAE and probabilistic generative models, e.g., GMM and HMM, Johnson et al. proposed a probabilistic generative model with VAEs as emission distributions~\cite{johnson2016composing}. Employing a neural network as a part of the probabilistic generative model and as a part of inference procedure, i.e., an inference network, is now broadening the possibility of applications of probabilistic generative models and neural networks. Edward, a probabilistic programming environment developed by Tran et al.~\cite{tran2016edward}, has already been merged into TensorFlow, which is the most popular deep learning framework. Finding an appropriate way to integrate probabilistic generative models and neuro-dynamics models is crucial for developing computational cognitive architecture modeling symbol emergence in cognitive developmental systems.}

\addspan{Artificial systems involving symbol emergence need to have many cognitive components, e.g., visual and speech recognition, motion planning, word discovery, speech synthesis, and logical inference. All of those must be adaptive and learn through interaction with the system's environment in a lifelong manner. In addition, when we develop such cognitive components, we need to integrate them into a conjoint architecture. After integration, the system should act as a single, comprehensive learning system. Developing an architecture that is decomposable and comprehensible while providing consistent learning results is important. Nakamura et al. introduced a framework called SERKET, which enable us to develop cognitive components independently and allows them to work and learn together in the same theoretical way, as they are developed as a single learning system~\cite{nakamura2018serket}. Computational models and frameworks to develop a large-scale cognitive architecture that can work practically in the real-world environment is our challenge.}

\subsection{Robotic planning with grounded symbols} 

Since the early days of AI, symbolic planning techniques have been employed to allow agents to achieve complex tasks in closed and deterministic worlds. As discussed previously, one problem in applying such symbolic reasoning to robotics is symbol grounding~\cite{coradeschi2013short}, i.e., mapping symbols to syntactic structures of concrete objects, paths, and events. 

In robotics, this problem has been rephrased using the term symbol anchoring, which puts stronger emphasis on the linking of a symbol (in symbolic AI) to real-world representations acquired with robot sensing~\cite{coradeschi:2003:ras,lemaignan:2012:ijsr,elfring:2013:ras}. This is a particularly challenging task when the process takes place over time in a dynamic environment. A number of works in the literature have attempted to solve this problem by relying on the concept of affordances~\cite{Gibson1979,Woergoetter2009,Krueger2011,jamone16aff,zech2017}: action possibilities that are perceived in the objects. The perception of object affordances relies on an action-centric representation of the object, which is dependent on the sensorimotor abilities of the agent and that is learned from experience: if a robot learns how to perceive affordances based on its own visuomotor experience, it will then be able to attach action-related symbolic information (e.g.,\ afforded action, effects of the action) to visual perceptions. 

In the AI and robotics literature, comprehensive cognitive architectures for representation, planning, and control in robots have been proposed, involving multiple components addressing different requirements (e.g., sensor fusion, inference, execution monitoring, and manipulation under uncertainty). However, there is currently no single framework that is suited for all applications given their great diversity~\cite{siciliano:handbook}. Examples of such comprehensive systems~\cite{py:2010:aamas,sisbot:2012:tro}, while presenting solid theoretical foundations in behavior-based control and robot simulation results, still lack robust and general applicability (e.g., not being restricted to one specific task) on real robot platforms, such as humanoids.

The above discussion shows that one major challenge arises from having to design systems that can be scaled up to complex problems, while all their components need to allow for grounding. At the same time, such systems must cope with the nasty contingencies and the noise of their real sensorimotor world. Recently, \emph{hybrid} approaches that combine strategies of symbolic AI-based planning and perception- and behavior-based robot control have been proposed to overcome the main limitations of the two philosophies~\cite{matuszek:2012:icml,matuszek:2013:er,chen_mooney:2011:aaai,pastra:2008:praxicon,dzifcak:2009:icra,antunes16planning}: whereas symbolic planning offers good generalization capabilities and the possibility to scale, its static and deterministic set of rules does not cope well with real uncertain environments where robots operate. Conversely, tight perception--action loops that allow robots to perform simple tasks effectively in the real world do not scale to more complex problem solving. 

From the viewpoint of symbol emergence system, action planning capability and syntax should also be organized in a bottom-up manner (Fig.~\ref{fig:multimodal_fusion}). Structural bootstrapping had been introduced as one possible model \cite{Woergoetter2015}, in which the structural similarity of a known to an unknown action is used as a scaffold to make inferences of how to ``do it with the new one." \addspan{Additionally, syntactic bootstrapping also becomes a key to accelerate learning action planning and language. If the internal representation of planning and syntax are shared in our cognitive system as shown in Fig.~\ref{fig:multimodal_fusion}, it will accelerate, i.e., bootstrap, syntax and planning capability learning.}

This relates to learning hierarchies of action concepts as well. When forming action concepts and symbols from sensorimotor experience, what should be their scale or granularity? If internal representations are very fine-grained (\emph{bend a finger}), they are easy for a robot to learn but hardly represent useful cause--effect relations and are of limited use for action planning. If they are very coarse-grained (\emph{build a tower}), then they represent powerful concepts but are very difficult to learn from sensorimotor interaction. One way to obtain the best of both worlds is to allow the agent to learn simple concepts from sensorimotor experience, and then learn more complex concepts on the basis of what it has already learned.

Ugur and Piater~\cite{Ugur2015, Ugur-2016-TCDS} illustrated this principle in the context of a robot that learns to build towers from experience by child-like play. The central idea behind this work was to allow a robot to first (pre-trial) learn basic concepts asking: ``How will individual tower-building blocks respond to actions?'' In doing so, the robot was also categorizing them by their properties discovered in this way. It could then be shown that learning the stackability of building blocks is clearly facilitated by using the category labels from the pre-trials as additional input features. Similarly, generative models can be beneficially used to learn and categorize actions in a hierarchical way.

\delspan{{\it C. Scaling Up}}

\delspan{Humans are seemingly able to learn concepts of virtually unlimited complexity. From a machine learning perspective, this is possible thanks to two important principles.  The first lies in \emph{learning new concepts in terms of already-learned concepts} such that no learning problem is very difficult, but the final result can achieve very high levels of complexity. For example, scalar multiplication is a simple extension of addition, which is in turn a simple generalization of counting.}

\delspan{The second principle is the acquisition of \emph{general knowledge} that can be exploited to facilitate future learning.}   

\delspan{General knowledge---like having learned that touching a hot surface causes pain--- massively  simplifies future learning problems because it allows the learner to exclude large swaths of the solution space, e.g.\ not having to ask whether \textit{any} hot surface does this.  The more such knowledge a learner acquires, the simpler future learning problems become.}

\delspan{The first principle was explored in robotics and AI for example by Ugur and Piater, but the question remains whether the concept of hierarchical learning can be scaled up to solve complex problems?  Alas, it must be noted that in these studies each partial learning problem was engineered to make learning possible.  The robot had only a small action and perception repertoire to limit the space of learnable concepts. Thus, in making the targeted task \textit{easy} to learn, this setup makes it \textit{hard} for the robot to learn large classes of other tasks.  This is a limitation of practically all current machine learning systems.}

\delspan{The second principle remains largely unexplored.  One challenge lies in the design of learning methods and data abstractions that allow \textit{generic} principles to be represented.  The same pre-structuring problem occurs here again: Learning requires appropriately-engineered data structures that make the relevant cues discoverable, but these necessarily constrain the scope of what can be learned. This is also known as the \textit{bias-variance dilemma}. If you put a lot of bias (pre-sturcturing) into your learner, it will learn its task fast but only this task. Without much bias it will be able to learn other tasks, too, increasing the variability (variance), but this will take long.}

\delspan{How do humans get around this problem?  The answer is: they do not. Human exploratory learning is faced with the same needle-in-a-haystack problem of extracting relevant predictors from very high-dimensional feature spaces using limited amounts of training data.}

\delspan{Crucially, however, humans do not only learn by exploration.  Humans conserve knowledge in symbolic form and pass it on to other humans, allowing others to take advantage of it without the need to rediscover it, and allowing knowledge to accumulate over generations.  Scaling up robot learning will require equivalent mechanisms.}

\subsection{Language acquisition by a robot}
Language acquisition is one of the biggest mysteries of the human cognitive developmental process. Asada et al.\ described how cognitive developmental robotics aims to provide new understanding of human high-level cognitive functions, and here specifically, also the language faculty, developed by means of a constructive approach that developmentally generates those cognitive functions~\cite{Asada2009}.

Along these lines, an integrative model that can simultaneously explain the process of language acquisition and the use of the language obtained is still missing. To understand the process of language acquisition, we need to develop a computational model that can capture the dynamic process of the language-learning capabilities, e.g., learning vocabulary, constructing mutual belief, forming multimodal concepts, learning syntax, and understanding metaphors. It appears that a computational model based on machine learning could provide a promising approach for understanding cognitive phenomena related to language acquisition.

A series of \addspan{non-developmental} studies using deep and recurrent neural networks made significant progress in speech recognition and natural language processing, e.g., machine translation, image captioning, word embedding, and visual question answering\addspan{, using a large amount of labeled data to create off-the-shelf pattern processing systems}. However, this approach has, at least, two drawbacks. First, the approach is heavily based on manually labeled data or prepared parallel corpora, e.g., transcribed speech data and a set of pairs of English and Japanese texts. The learning process is clearly different from that of human infants. Human infants never have access to transcribed speech signals when they acquire phonemes, words, and syntax. Human infants learn language using their multimodal sensorimotor experiences. Thus, current computational models that are based on supervised learning \addspan{are basically not suitable for} a constructive model of human child cognitive development. Second, semantic understanding of language by an AI system is inevitably limited without a grounded internal representation system (e.g.,\ in terms of sensorimotor PSS). For further improvement in the performance of translation, dialogue, and language understanding, a computational model that includes a comprehensive language-learning process will be required. 

Recent advances in modern AI and robotics have brought the possibility of creating embodied computational intelligence that behaves adaptively in a real-world environment.
Creating a robot that can learn language from its own sensorimotor experience alone is one of our challenges, which is an essential element for the understanding of symbol emergence in cognitive systems. \addspan{Many studies have been exploring the challenge in modeling language acquisition in developmental process using neural networks~\cite{tani2016exploring,heinrich2018interactive,antunes2017communication} and probabilistic models~\cite{Akira2017,Akira_angelo,nishihara2017online,Taniguchi2016SER}.}

There is quite lengthy list of subchallenges that have to be addressed before we can reach this goal. The following enumeration is incomplete.
\begin{enumerate}
\item
A developmental robot should acquire phonemes and words from speech signals directly without transcribed data. Some attempts toward this exist. For example, Taniguchi and coworkers proposed a hierarchical Dirichlet process-based hidden language model, which is a nonparametric Bayesian generative model, and showed that the method lets a robot acquire words and phonemes from speech signals alone. However, it still has many restrictions~\cite{Taniguchi2016,Taniguchi2016b}. 

\item
Learning the meaning of words and phrases based on multimodal concept formation is another subchallenge. The interplay between different modalities is crucially important for this. For example, the learning process of the meaning of a verb, e.g., ``hit'' or ``throw,'' must depend on the learning process of the motion primitives, action concepts, and perceptual outcomes of these actions.

\item
We also need a machine learning method for syntax learning. Reliable unsupervised syntax learning methods are still missing. Many scientists believe that the syntactic structure in action planning can be used for bootstrapping language syntax learning. 

\item
Furthermore, we need unsupervised machine learning methods for the estimation of speech acts and meaning of function words, e.g., prepositions, determiners, and pronouns, as well as metaphors without artificially prepared labeled data.

\item
The learning of nonlinguistic symbol systems that are used with linguistic expressions, e.g., gestures, gaze, and pointing, is also important. They might even have to be learned as prerequisites of language learning~\cite{tomasello2010}.
\end{enumerate}

All these subchallenges, in addition, need to deal with the uncertainty and noise in speech signals and observed (sensorimotor) events.

\subsection{Human--robot communication, situated symbols, and mutual beliefs} 
The existence of shared symbol systems is clearly fundamental in human--robot collaboration. Cognitive artificial systems that incorporate the translation of natural-language instructions into robot knowledge and actions have been proposed in many studies. For example, the work by Tellex et al.\ is geared toward interpreting language commands given to mobile robots~\cite{tellex:2011:aaai} and toward statistical symbol grounding~\cite{tellex:2011:ai}.

In order to support human activities in everyday life, robots should adapt their behavior according to the situations, which differ from user to user. In order to realize such adaptation, these robots should have the ability to share experiences with humans in the physical world. Moreover, this ability should also be considered in terms of spoken language communication. The process of human language communication is based on certain beliefs shared by those communicating with each other~\cite{sperber1995}. From this viewpoint, language systems can be considered as shared belief systems that include not only linguistic but also nonlinguistic shared beliefs, where the latter are used to convey meanings based on their relevance for the linguistic beliefs.

However, the language models existing in robotics so far are characterized by fixed linguistic knowledge and do not make this possible~\cite{allen2001toward}. In these methods, information is represented and processed by symbols, the meaning of which have been predefined by the robots' developers. Therefore, experiences shared by a user and a robot do not exist and can, thus, neither be expressed nor interpreted. As a result, users and robots fail to interact in a way that accurately reflects human--human communication. 

To overcome this problem and to achieve natural dialogue between humans and robots, we should use methods that make it possible for humans and robots to share symbols and beliefs based on shared experiences. To form such shared beliefs, the robot should possess a mechanism, related to mechanisms of the \emph{theory of mind}, that enables the human and the robot to infer the state of each other's beliefs, better allowing them to coordinate their utterances and actions. Iwahashi et al.~\cite{iwahashi2007robots, iwahashi2010robots} presented a language acquisition method that begins to allow such inference and makes it possible for the human and the robot to share at least some beliefs for multimodal communication.

\delspan{{\it F. Lifelong learning}}

\delspan{Traditional AI starts with amodal symbol systems. Thus, symbols, i.e, discrete internal representations, were typically designed up front, and their grounding remained an issue. Instead, in humans, we see that the emergence of symbols is a result of a bottom-up, clustering-like process intertwined with the influence of language and communication with others and with physical sensorimotor experiences. This process is more plastic early in development, but never really stops. In artificial systems, this has not yet been paralleled: learning is typically restricted to very brief periods (hours, typically) and lifelong learning is still a distant goal. Greater reliability and robustness of robots is a key pre-condition for this.}

\delspan{Artificial systems involving symbol emergence need to have many cognitive components (sensor as well as motor). All of those must be adaptive and learn through interaction with the system's environment in a lifelong manner. In addition, when we develop such cognitive components, we need to integrate them into a conjoint architecture. After integration, the system should act as a single, comprehensive learning system. Developing an architecture that is decomposable and comprehensible while providing consistent, life-long learning capability is a massive challenge in this field.}

\section{Conclusion}\label{sec:Conclusion}
 
AI, cognitive science, neuroscience, developmental psychology, and machine learning have been discussing \emph{symbols} in different contexts, but usually sharing similar ideas to a certain extent. However, these discussions in the different fields have been causing various types of confusion. A central problem here is still the mixing up of the aspect of symbol systems in society with the aspect of a rather agent-centric, internal representation-based symbol system. Here, we have tried to point out the importance of \textit{symbol emergence} in cognitive developmental systems by the interaction of both aspects. Another aspect that we deem important is that symbol emergence should be posed as a multifaceted problem that should take different sources of information for learning and different points for exploitation into account. To help in this process, we tried to more clearly define and reformulate different terms from the perspective of (artificial) developmental cognitive systems. This may, thus, represent a biased view, but we think that it might benefit our field. 

Machine learning-based AI, including deep learning, has recently been considered as a central topic in the AI community. In this context, it is often said that the notion of a ``symbol system'' is out of date.
\delspan{ and \emph{everything needed} is encoded implicitly in the weights of such networks However, these ``modern views'' appear, when considered from a different perspective, rather old-fashioned themselves, as the above statement is implying an amodal view onto symbol systems (ASS).}
\addspan{However, the ``symbol system'' refers to symbolic AI, and not a symbol system in our society. We humans still uses symbols, e.g., language, to think and communicate. Learning and using symbols is still a challenge in AI studies. To go further, we needed to disentangle the two different notions of symbols: symbol systems in symbolic AI and those in our human society. The symbol emergence problem is not about symbolic AI, but rather about cognitive developmental systems dealing with (emergent) symbol systems in our human society.} 

Clearly, studying the interaction between language and internal representations in the context of the more fluent, dynamic, and possibly more brain-like PSS is not out of date. We also believe that symbol emergence is not an auxiliary problem in cognitive developmental science, including cognitive developmental robotics and AI, but central to understanding and developing natural and artificial cognitive systems. Thus, studying symbol emergence in natural and artificial cognitive systems, and tackling the problems described in Section~\ref{sec:Challenges} remain essential future challenges.

\bibliographystyle{IEEEtran}
\bibliography{symbol_emergence}

\begin{IEEEbiography}[{\includegraphics[width=1in,height=1.25in,clip,keepaspectratio]{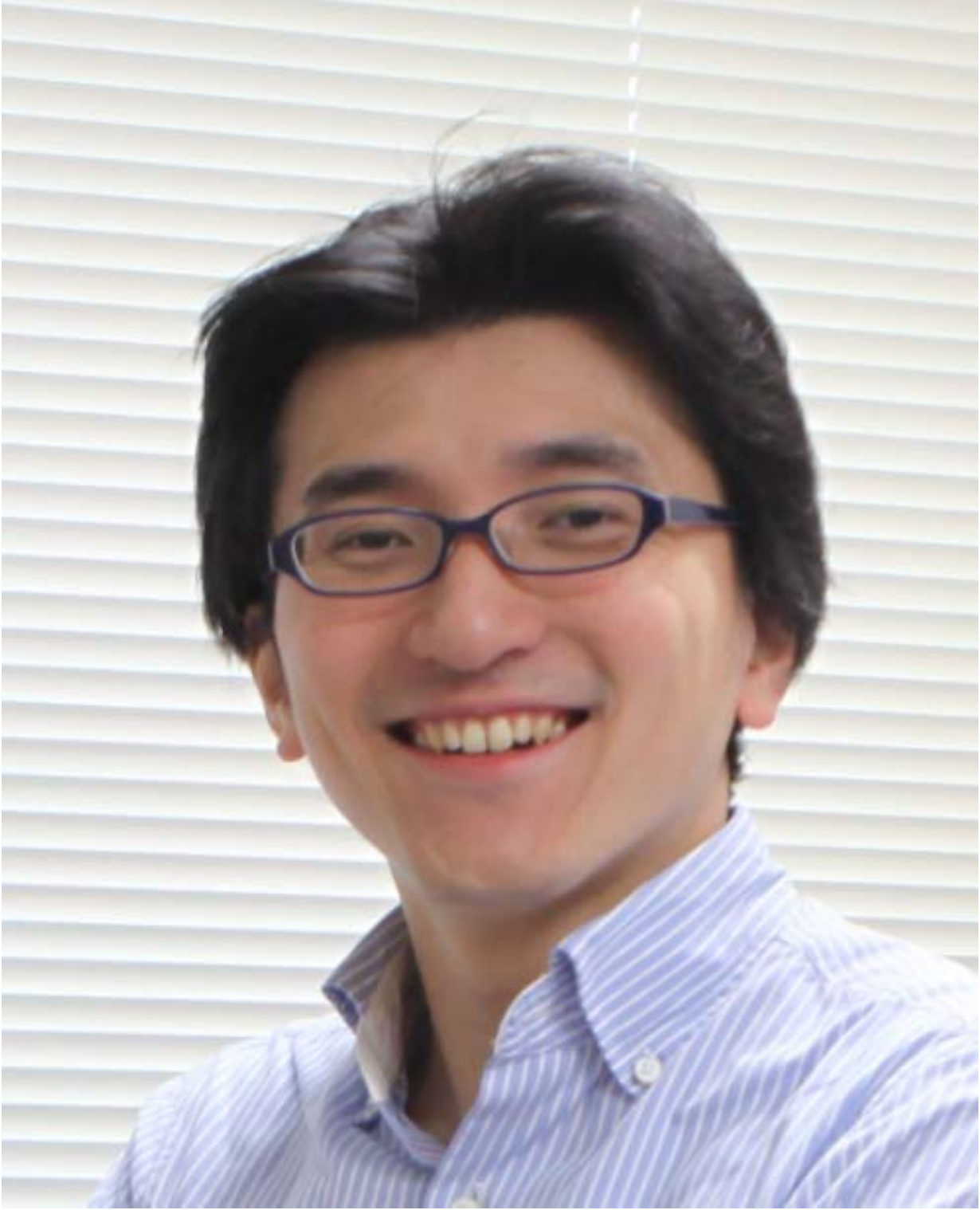}}]
{Tadahiro~Taniguchi} received the ME and PhD degrees from Kyoto University in 2003 and 2006, respectively. From April 2005 to March 2006, he was a Japan Society for the Promotion of Science (JSPS) research fellow (DC2) in the Department of Mechanical Engineering and Science, Graduate School of Engineering, Kyoto University. From April 2006 to March 2007, he was a JSPS research fellow (PD) in the same department. From April 2007 to March 2008, he was a JSPS research fellow in the Department of Systems Science, Graduate School of Informatics, Kyoto University. From April 2008 to March 2010, he was an Assistant Professor at the Department of Human and Computer Intelligence, Ritsumeikan University. Since April 2010, he has been an associate professor in the same department. He has been engaged in research on machine learning, emergent systems, and semiotics.
\end{IEEEbiography}

\begin{IEEEbiography}[{\includegraphics[width=1in,height=1.25in,clip,keepaspectratio]{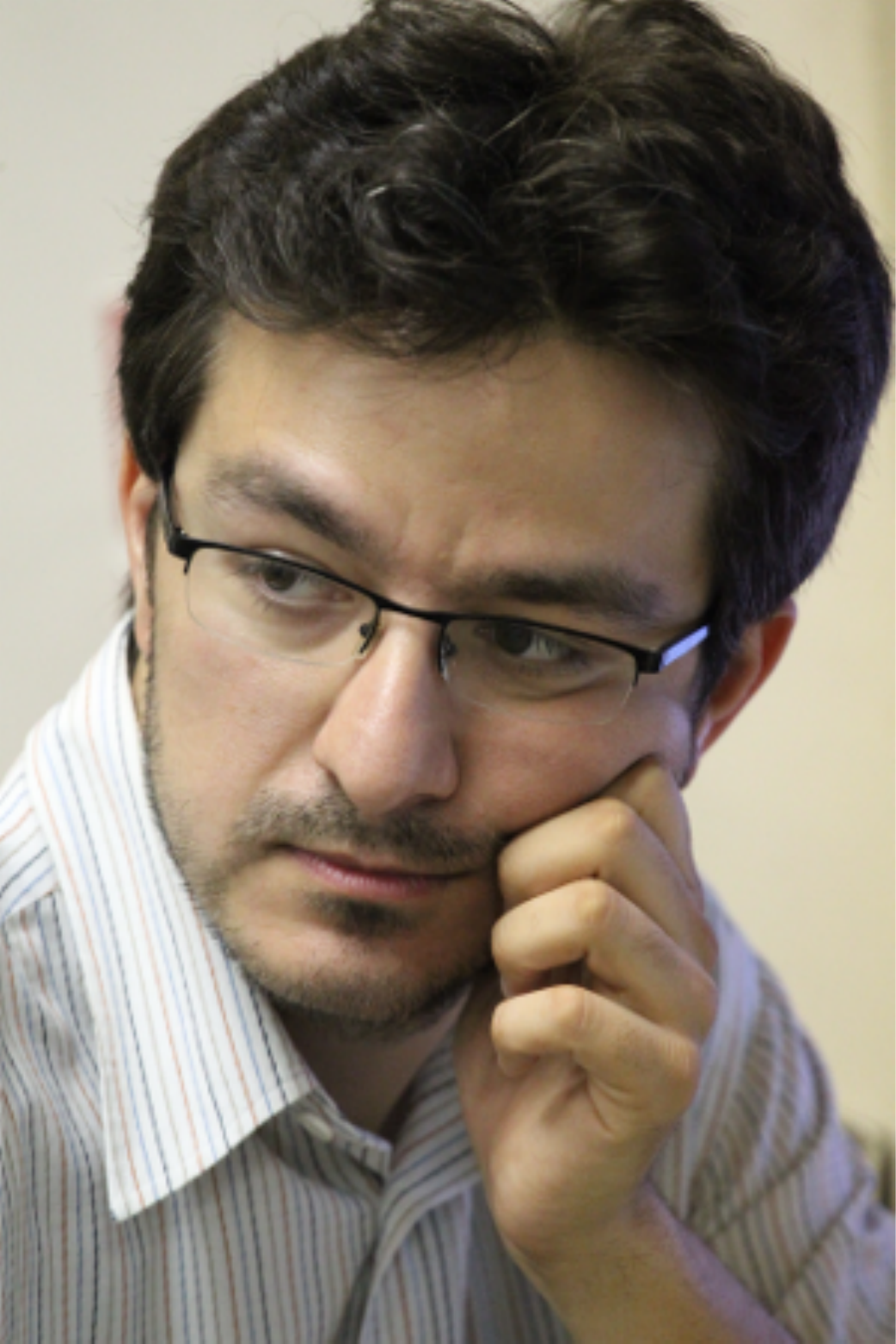}}]
{Emre~Ugur } is an Assistant Professor in Department of Computer Engineering Department, Bogazici University, Turkey. After receiving his PhD in Computer Engineering from Middle East Technical University, he worked at ATR Japan as a researcher (2009--2013), at University of Innsbruck as a senior researcher (2013--2016), and at Osaka University as a specially appointed Assistant Professor (2015--2016). He is interested in developmental and cognitive robotics, intelligent manipulation, affordances, and brain--robot interface methods.

\end{IEEEbiography}

\begin{IEEEbiography}[{\includegraphics[width=1in,height=1.25in,clip,keepaspectratio]{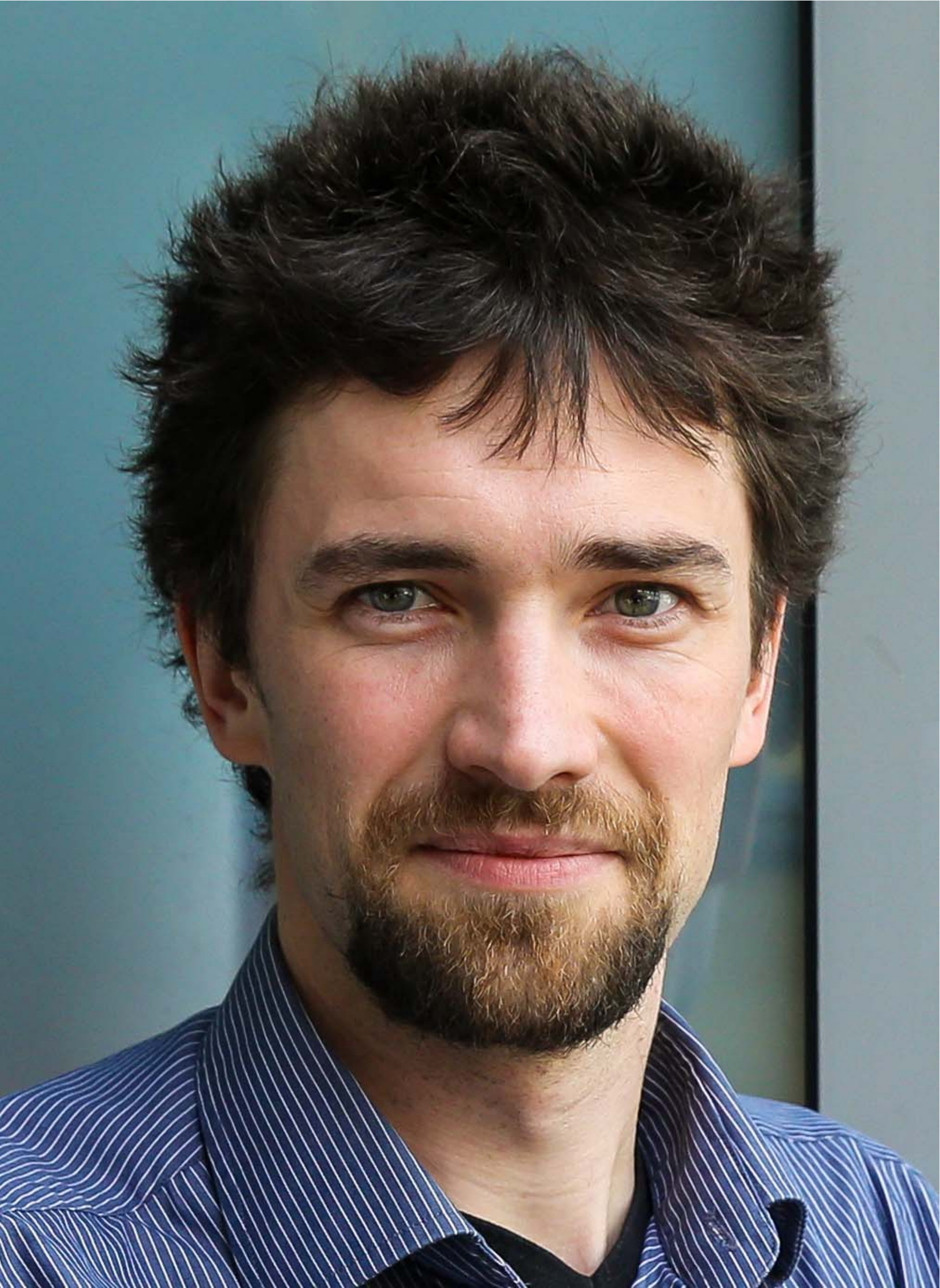}}]
{Matej ~Hoffmann} received the Mgr. (MSc) degree in computer science, artificial intelligence at Faculty of Mathematics and Physics, Charles University in Prague, Prague, Czech Republic, in 2006, and the PhD degree from the Artificial Intelligence Laboratory, University of Zurich, Zurich, Switzerland, under the supervision of Prof. Rolf Pfeifer. 

He then served as a Senior Research Associate with the Artificial Intelligence Laboratory, University of Zurich, from 2012 to 2013. From 2013 to 2017, he was with the iCub Facility Department, Istituto Italiano di Tecnologia (IIT), Genoa, Italy, where he was a Marie Curie Experienced Researcher Fellow, from 2014 to 2016. In 2017 he joined the Department of Cybernetics, Faculty of Electrical Engineering, Czech Technical University in Prague. His current research interests include embodied cognition and developmental robotics, in particular the mechanisms underlying body representations and sensorimotor contingencies in humans and their implications for increasing the autonomy, resilience, and robustness of robots.

\end{IEEEbiography}

\begin{IEEEbiography}[{\includegraphics[width=1in,height=1.25in,clip,keepaspectratio]{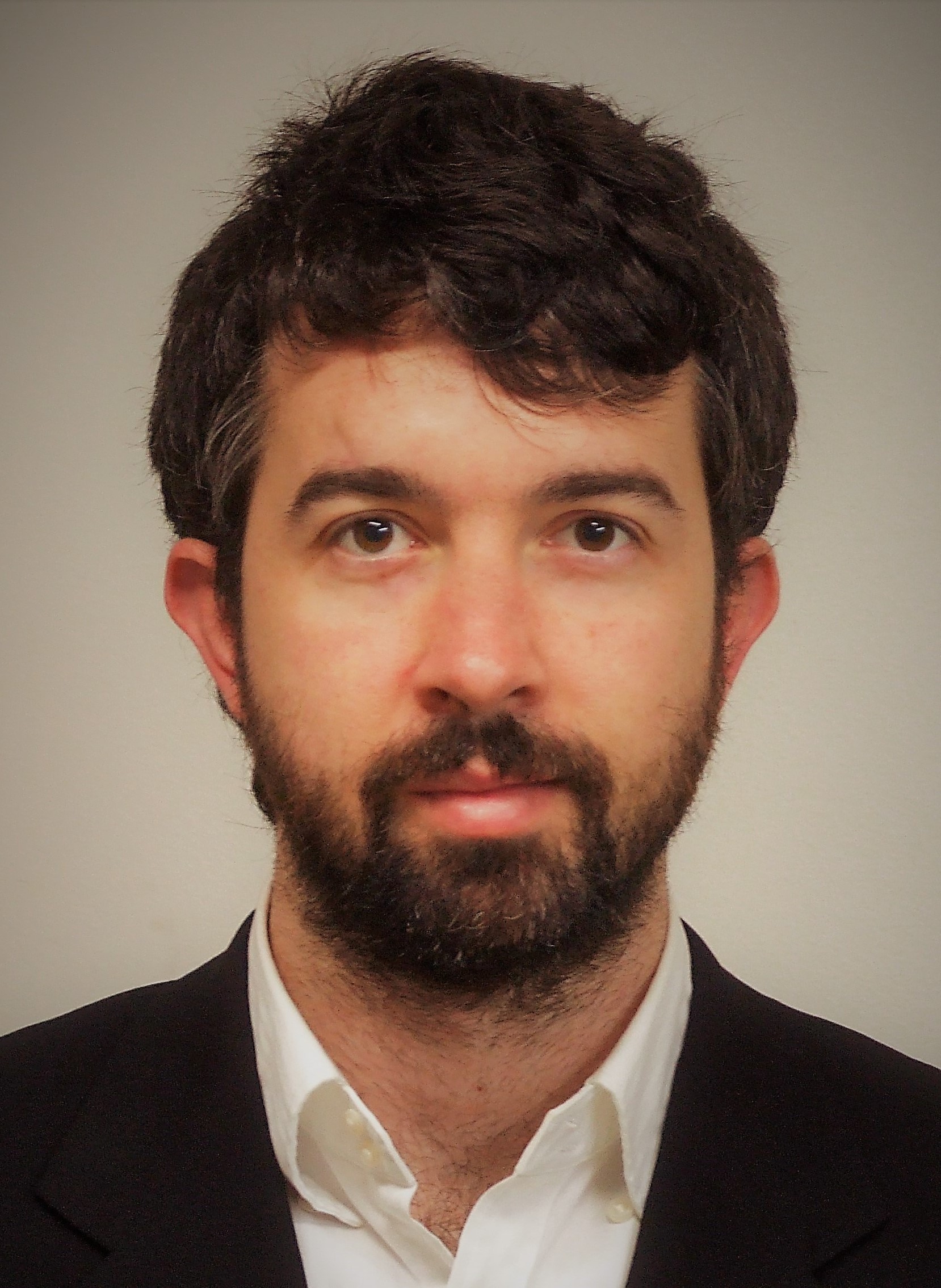}}]
{Lorenzo~Jamone} received his MS in Computer Engineering from the University of Genova in 2006 (with honors), and his PhD in Humanoid Technologies from the University of Genova and the IIT in 2010. He was a Research Fellow at the RBCS Department of the IIT in 2010, Associate Researcher at Takanishi Lab (Waseda University, Tokyo, Japan) from 2010 to 2012, and Associate Researcher at VisLab (the Computer Vision Laboratory of the Instituto de Sistemas e Robotica, Instituto Superior T\'ecnico, Lisbon, Portugal) from 2013 to 2016. Since 2016, he has been a Lecturer in Robotics at the Queen Mary University of London. His main research interests include: sensorimotor learning and control in humans and robots; robotic reaching, grasping, manipulation, and tool use; force and tactile sensing; cognitive developmental robotics.
\end{IEEEbiography}

\begin{IEEEbiography}[{\includegraphics[width=1in,height=1.25in,clip,keepaspectratio]{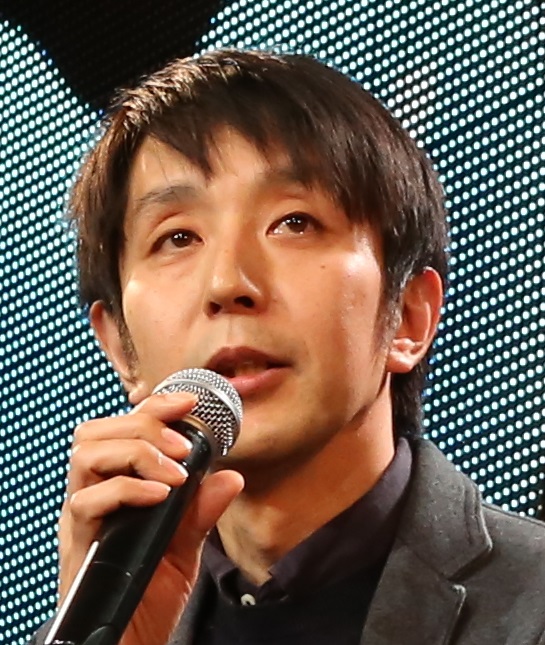}}]
{Takayuki~Nagai} received his BE, ME, and PhD degrees from the Department of Electrical Engineering, Keio University, in 1993, 1995, and 1997, respectively. Since 1998, he has been with the University of Electro-Communications, where he is currently a professor of the Graduate School of Informatics and Engineering. From 2002 to 2003, he was a visiting scholar at the Department of Electrical Computer Engineering, University of California, San Diego. Since 2011, he has also been a visiting researcher at Tamagawa University Brain Science Institute. He has received the 2013 Advanced Robotics Best Paper Award. His research areas include intelligent robotics, cognitive developmental robotics, and human--robot interaction. He is a member of the IEEE, RSJ, JSAI, IEICE, and IPSJ.
\end{IEEEbiography}

\begin{IEEEbiography}[{\includegraphics[width=1in,height=1.25in,clip,keepaspectratio]{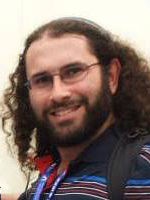}}]
{Benjamin~Rosman} is a Principal Researcher in the Mobile Intelligent Autonomous Systems group at the Council for Scientific and Industrial Research (CSIR) in South Africa, and is also a Senior Lecturer in the School of Computer Science and Applied Mathematics at the University of the Witwatersrand, where he runs the Robotics, Autonomous Intelligence and Learning Laboratory. He received his PhD degree in Informatics from the University of Edinburgh in 2014, and previously obtained his MSc in Artificial Intelligence from the University of Edinburgh. His research interests focus on learning and decision making in autonomous systems, in particular how learning can be accelerated through generalizing knowledge from related problems. He also works in skill learning and transfer for robots. He currently serves as the Chair of the IEEE South Africa Joint Chapter of Control Systems \& Robotics and Automation.
\end{IEEEbiography}

\begin{IEEEbiography}[{\includegraphics[width=1in,height=1.25in,clip,keepaspectratio]{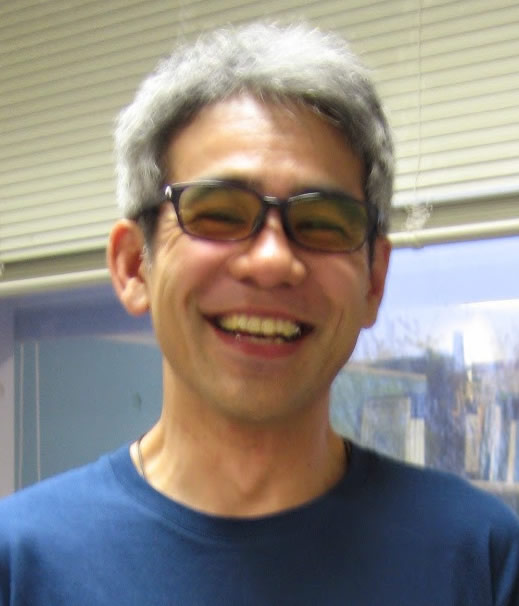}}]
{Toshihiko~Matsuka} received his PhD in Psychometrics from Columbia University in 2002. After receiving his PhD, he joined RUMBA lab at Rutgers University, Newark as a postdoctoral research fellow. While at Rutgers, he conducted research on cognitive neuroscience. He then moved to Stevens Institute of Technology as an Assistant Research Professor, conducting research on Information Science. Since 2007 he has been with Chiba University, where he is currently a Professor in the Department of Cognitive Information Science. His research interests include: category/concept learning, decision making, other high-order cognitive processing, social cognition, and computational cognitive modeling. 

\end{IEEEbiography}

\begin{IEEEbiography}[{\includegraphics[width=1in,height=1.25in,clip,keepaspectratio]{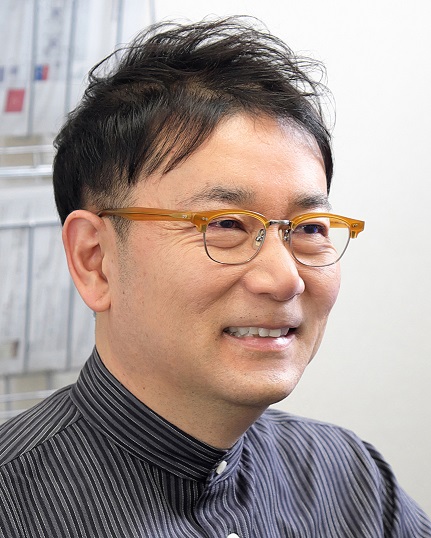}}]
{Naoto~Iwahashi} received the BE degree in Engineering from Keio University, Yokohama, Japan, in 1985. He received the PhD degree in Engineering from Tokyo Institute of Technology, in 2001. In April 1985, he joined Sony Corp., Tokyo, Japan. From October 1990 to September 1993, he was at Advanced Telecommunications Research Institute International (ATR), Kyoto, Japan. From October 1998 to June 2004, he was with Sony Computer Science Laboratories Inc., Tokyo, Japan. From July 2004 to March 2010, he was with ATR. From November 2005 to March 2011, he was a visiting professor at Kobe University. In April 2008, he joined the National Institute of Information and Communications Technology, Kyoto, Japan. Since April 2014, he has been a professor at Okayama Prefectural University. Since April 2011, he has also been a visiting researcher at Tamagawa University Brain Science Institute. His research areas include machine learning, artificial intelligence, spoken language processing, human--robot interaction, developmental multimodal dialog systems, and language acquisition robots.
\end{IEEEbiography}

\begin{IEEEbiography}[{\includegraphics[width=1in,height=1.25in,clip,keepaspectratio]{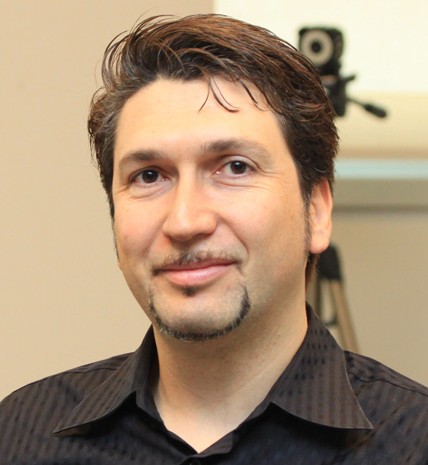}}]
{Erhan~Oztop} earned his PhD at the University of Southern California in 2002. In the same year, he joined the Computational Neuroscience Laboratories at the Advanced Telecommunications Research Institute International (ATR), in Japan. There, he worked as a researcher and later a senior researcher and group leader, in addition to serving as vice department head for two research groups and holding a visiting associate professor position at Osaka University (2002--2011). Currently, he is a faculty member and the chair of the Computer Science Department at Ozyegin University, Istanbul. His research involves computational study of intelligent behavior, human-in-the loop systems, computational neuroscience, machine learning, and cognitive and developmental robotics.
\end{IEEEbiography}

\begin{IEEEbiography}[{\includegraphics[width=1in,height=1.25in,clip,keepaspectratio]{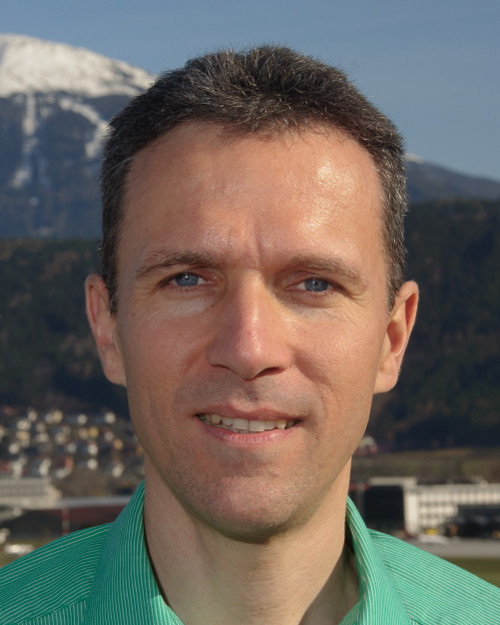}}]
{Justus~Piater} is a professor of computer science at the University of
Innsbruck, Austria, where he leads the Intelligent and Interactive
Systems group. He holds a MSc degree from the University of
Magdeburg, Germany, and MSc and PhD degrees from the University of
Massachusetts Amherst, USA, all in computer science. Before joining
the University of Innsbruck in 2010, he was a visiting researcher at
the Max Planck Institute for Biological Cybernetics in T\"ubingen,
Germany, a professor of computer science at the University of Li\`ege,
Belgium, and a Marie-Curie research fellow at GRAVIR-IMAG, INRIA
Rh\^one-Alpes, France.  His research interests focus on visual
perception, learning, and inference in sensorimotor systems. He has
published more than 160 papers in international journals and
conferences, several of which have received best-paper awards, and
currently serves as Associate Editor of the IEEE Transactions on
Robotics.
\end{IEEEbiography}

\begin{IEEEbiography}[{\includegraphics[width=1in,height=1.25in,clip,keepaspectratio]{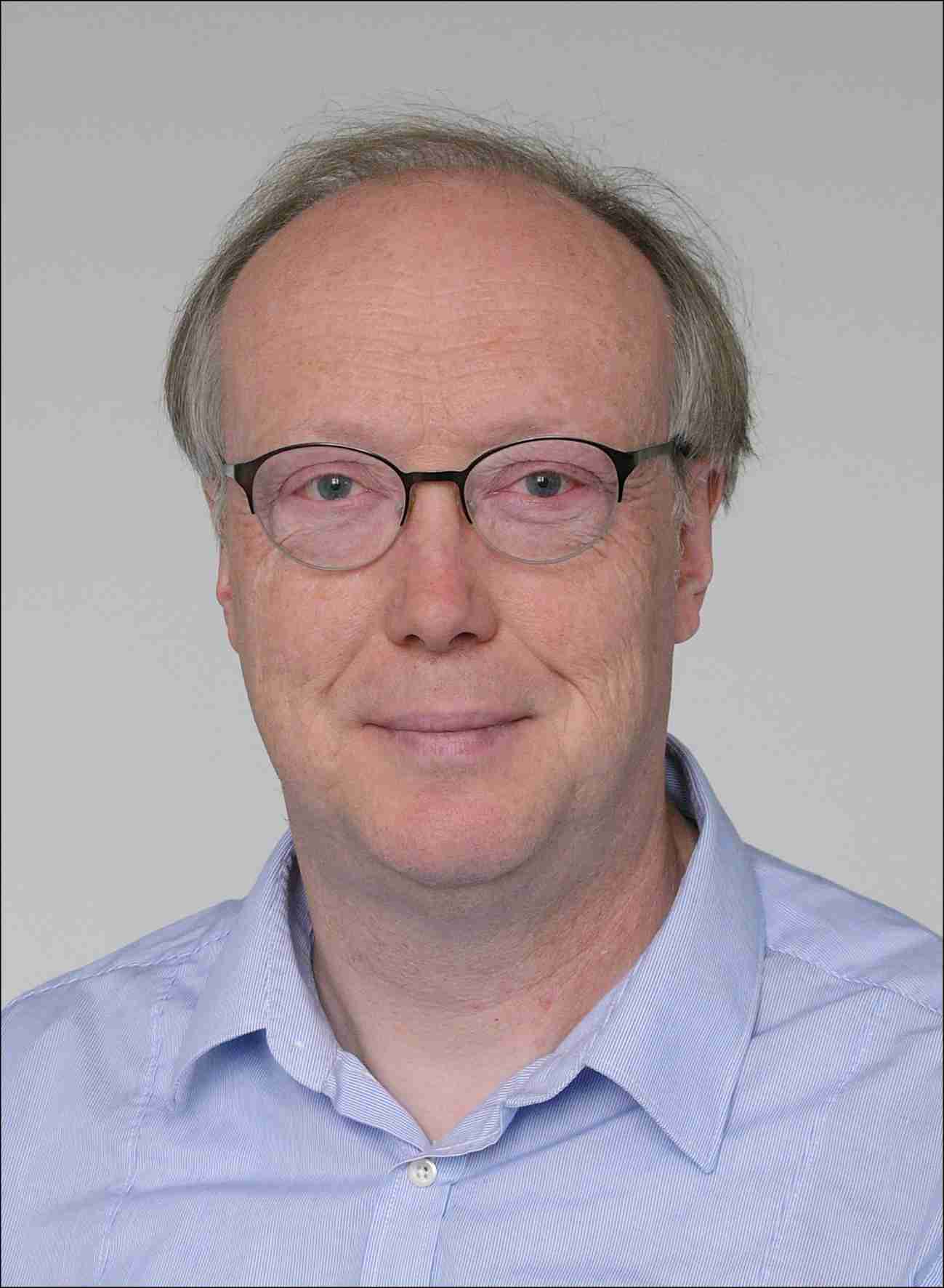}}]
{Florentin~W\"org\"otter} has studied Biology and Mathematics in D\"usseldorf. He received his PhD in 1988 in Essen, working experimentally on the visual cortex before he turned to computational issues at Caltech, USA (1988--1990). After 1990, he was a researcher at the University of Bochum, concerned with experimental and computational neuroscience of the visual system. Between 2000 and 2005, he was a Professor for Computational Neuroscience at the Psychology Department of the University of Stirling, Scotland where his interests strongly turned toward \emph{learning in neurons}. Since July 2005, he has led the Department for Computational Neuroscience at the Bernstein Center at the University of G\"ottingen, Physics III. His main research interest is information processing in closed-loop perception--action systems, which includes aspects of sensory processing (vision), motor control, and learning/plasticity. These approaches are tested in walking as well as other robotic implementations. His group has developed the RunBot a fast and adaptive biped walking robot based on neural control. Such neural technologies are currently also being investigated in the context of advanced ortheses.
\end{IEEEbiography}	
\end{document}